\documentclass[twoside,11pt]{article}

\usepackage{jmlr2e}
\usepackage{amsmath,amssymb,mathrsfs}
\usepackage{latexsym,amsfonts,amscd,amsxtra,amstext}
\usepackage{algorithm,algorithmic}
\usepackage{url}
\usepackage{bm}
\usepackage{xcolor}
\definecolor{babyblack}{RGB}{246, 218, 101} 
\definecolor{springgreen}{RGB}{255, 255, 133} 
\definecolor{orchid}{RGB}{93, 173, 226} 
\definecolor{limegreen}{RGB}{145, 223, 208} 
\usepackage{multirow}
\usepackage{wrapfig}
\usepackage{verbatim}
\usepackage{caption}
\usepackage{subcaption}
\usepackage{graphicx}
\usepackage{makecell}
\usepackage{mathtools}
\usepackage{xr}
\usepackage{booktabs}
\usepackage{colortbl}
\usepackage{tikz}
\usetikzlibrary{shapes.geometric, arrows, positioning}

\newcommand{\RR}{\mathbb{R}}

\newcommand{\EE}{\mathbb{E}}

\newcommand{\Gc}{\mathbb{G}}
\newcommand{\Ch}{\mathrm{Clip}}
\newtheorem{assumption}{Assumption}
\usepackage{epstopdf}

\usepackage{lastpage}
\jmlrheading{26}{2025}{1-\pageref{LastPage}}{11/24; Revised
9/25}{11/25}{24-1991}{Tao Sun,  Xinwang Liu, and Kun Yuan}
\ShortHeadings{Revisiting Gradient Control in Nonconvex SGD under Heavy-Tailed Noise}{T. Sun, X. Liu, and K. Yuan}

\firstpageno{1}

\begin{document}
	
	\title{
Revisiting Gradient Normalization and Clipping for Nonconvex SGD under Heavy-Tailed Noise: Necessity, Sufficiency, and Acceleration}
	
	\author{\name Tao Sun \email suntao.saltfish@outlook.com \\
		\addr College of Computer Science and Technology\\
		National University of Defense Technology\\
		Changsha, Hunan,  China
		\AND
		\name Xinwang Liu\thanks{Corresponding authors.} \email xinwangliu@nudt.edu.cn\\
		\addr College of Computer Science and Technology\\
		National University of Defense Technology\\
		Changsha, Hunan,  China
		\AND
		\name Kun Yuan$^\ast$\email kunyuan@pku.edu.cn\\
		\addr Center for Machine Learning Research (CMLR)\\
		Peking University\\
		Beijing, China
	}
	
\editor{Quanquan Gu}
	
	\maketitle

	\begin{abstract}
Gradient clipping has long been considered essential for ensuring the convergence of Stochastic Gradient Descent (SGD) in the presence of heavy-tailed gradient noise. In this paper, we revisit this belief and explore whether gradient normalization can serve as an effective alternative or complement. We prove that, under  individual smoothness assumptions, gradient normalization alone is sufficient to guarantee convergence of
 the nonconvex SGD. Moreover, when combined with clipping, it yields far better rates of convergence under   more challenging noise distributions. We provide a unifying theory describing normalization-only, clipping-only, and combined approaches. Moving forward, we investigate existing variance-reduced algorithms, establishing that, in such a setting, normalization alone is sufficient for convergence. Finally, we present an accelerated variant that under second-order smoothness improves convergence.
Our results provide theoretical insights and practical guidance for using normalization and clipping in nonconvex optimization with heavy-tailed noise.
	\end{abstract}
	
	\section{Introduction}
	Stochastic Gradient Descent (SGD) \citep{robbins1951stochastic}, a widely used method, serves as the primary optimization algorithm for addressing fundamental optimization problems in machine learning and statistics
	\begin{align}\label{model}
		\min_{\bm{w}\in\RR^d} f(\bm{w}):=\EE_{\xi\sim\mathcal{D}} f(\bm{w};\xi),
	\end{align}
	where $\mathcal{D}$ denotes the probability distribution over the statistical sample space $\Xi$. The convergence properties of SGD have been a significant area of research in machine learning, with many previous studies making substantial contributions to this topic \citep{ghadimi2013stochastic,cutkosky2019momentum,arjevani2023lower,johnson2013accelerating}. Early analyses often assumed that the noise caused by the stochastic gradient, denoted as $\bm{g}^t$, in each iteration could be uniformly variance-bounded, i.e., $\EE\|\bm{g}^t-\nabla f(\bm{w}^t)\|^2\leq \sigma^2$ for some $\sigma>0$ and all $t\in \mathbb{Z}^+$. However,   recent studies \citep{zhang2020adaptive,nguyen2019first,simsekli2019tail} have shown that this assumption of uniformly bounded variance is unrealistic for training neural networks, particularly in the context of language models \citep{kunstner2024heavy}.
	To address this, the concept of heavy-tailed noise is mathematically defined and utilized to analyze the convergence of SGD \citep{zhang2020adaptive}. Below, we revisit the assumption related to heavy-tailed noise\footnote{Indeed, there is an alternative definition of heavy-tailed noise. In \citep{gorbunov2020stochastic}, heavy-tailed noise is described as satisfying the uniformly bounded variance condition but not exhibiting sub-Gaussian behavior. This definition is, in fact, stricter than the assumption of bounded variance noise and even more so compared to Assumption \ref{ass2}. Our paper refers to the heavy-tailed noise as Assumption \ref{ass2}.}. 
	\begin{assumption}\label{ass2}
		There exists a constant $\sigma>0$  and $p\in (1,2]$ such that 
		$$
		\sup_{\bm{w}\in\RR^d}\{\EE_{\xi\sim \mathcal{D}}\|\nabla f(\bm{w};\xi)-\nabla f(\bm{w})\|^p\}\leq \sigma^p.
		$$
	\end{assumption}
	This assumption is significantly weaker than the standard bounded variance assumption. 	Specifically, when $p = 2$, this assumption reduces to the standard bounded variance assumption.
	At first glance, based solely on its mathematical form, Assumption \ref{ass2} might seem to be merely a minor relaxation of the standard bounded variance assumption, leading one to believe that it would have minimal impact on the analysis and performance of SGD. However, this assumption introduces a significant shift in the convergence properties of SGD. Specifically, in \citep{zhang2020adaptive}, the authors provide an example illustrating that when 
	$1<p<2$, SGD fails to converge, highlighting the critical implications of this seemingly modest change.

	\subsection{Main Approaches to Handling Heavy-Tailed Noise}\label{sec:main-approaches}
	To ensure the convergence of SGD in heavy-tailed noise, \citep{zhang2020adaptive} introduced a clipping technique, which they proved to be effective for handling such noise. The resulting method, SGDC, effectively reduces the impact of extreme gradient values, thereby stabilizing the training process. {The scheme of SGD with clipping is presented in Algorithm \ref{alg0}, where the clipping is defined as
		$\Ch_{h}(\bm{w}):=\min\Big\{1,\frac{h}{\|\bm{w}\|}\Big\}\bm{w}$
		with $\bm{w}\in\RR^d$.
		\begin{algorithm}[H]
			\caption{SGD with Clipping (SGDC) \citep{zhang2020adaptive}}\label{alg0}
			\begin{algorithmic}[1]
				\REQUIRE   parameters $\gamma>0$,  $h>0$\\
				\textbf{Initialization}: $\bm{w}^0$\\
				\textbf{for}~$t=0, 1,2,\ldots$ \\
				~~\textbf{step 1}: Samples unbiased stochastic gradient $\bm{g}^t$    \\
				~~\textbf{step 2}: Updates $ \bm{w}^{t+1}= \bm{w}^{t}-\gamma \Ch_{h}(\bm{g}^t)$ \\
				\textbf{end for}\\
			\end{algorithmic}
			\label{alg-SGDC}
		\end{algorithm}
	}
	
	Building on SGDC, \citep{cutkosky2021high} integrated gradient normalization with clipping, further enhancing the robustness of SGD against heavy-tailed noise in stochastic optimization. The details of this combined approach are presented in Algorithm \ref{alg1-c}. 
	\begin{algorithm}[H]
		\caption{Normalized SGD with Clipping (NSGDC) \citep{cutkosky2021high}}\label{alg1-c}
		\begin{algorithmic}[1]
			\REQUIRE   parameters $\gamma>0$,  $0\leq \theta<1$, $h>0$\\
			\textbf{Initialization}: $\bm{w}^0=\bm{w}^1$, $\bm{m}^0=\bm{0}$\\
			\textbf{for}~$t=1,2,\ldots$ \\
			~~\textbf{step 1}: Samples unbiased stochastic gradient $\bm{g}^t$    \\
			~~\textbf{step 2}: Computes $\bm{m}^t=\theta\bm{m}^{t-1}+(1-\theta)\Ch_{h}(\bm{g}^t)$\\
			~~\textbf{step 3}: Updates $ \bm{w}^{t+1}= \bm{w}^{t}-\gamma \frac{\bm{m}^t}{\|\bm{m}^t\|}$ \\
			\textbf{end for}\\
		\end{algorithmic}
		\label{alg-NSGDC}
	\end{algorithm}
	It has been proven that NSGDC converges under a very general smoothness assumption, i.e., the global gradient $\nabla f$ is Lipschitz. Additionally, when the gradient of the individual function $f(\cdot;\xi)$  is Lipschitz, a variance-reduced variant of NSGDC is established as discussed by \citet{liu2023breaking}, which is presented by Algorithm \ref{alg2-c}.
	\begin{algorithm}[H]
		\caption{Variance-Reduced NSGDC (NSGCD-VR) \citep{liu2023breaking}}\label{alg2-c}
		\begin{algorithmic}[1]
			\REQUIRE   parameters $\gamma>0$,  $0\leq \theta<1$, $h>0$\\
			\textbf{Initialization}: $\bm{w}^0=\bm{w}^1$, $\bm{m}^0=\bm{0}$\\
			\textbf{for}~$t=1,2,\ldots$ \\
			~~\textbf{step 1}: Sample the data  $\xi^t\sim\mathcal{D}$ and $\bm{g}^t=\nabla f(\bm{w}^t;\xi^t)$\\
			~~\textbf{step 2}:$\bm{m}^t=\theta\bm{m}^{t-1}+(1-\theta)\Ch_{h}(\bm{g}^t)+\theta\nabla f(\bm{w}^t;\xi^t)-\theta\nabla f(\bm{w}^{t-1};\xi^t)$\\
			~~\textbf{step 3}: $ \bm{w}^{t+1}= \bm{w}^{t}-\gamma \frac{\bm{m}^t}{\|\bm{m}^t\|}$ \\
			\textbf{end for}\\
		\end{algorithmic}
	\end{algorithm}
	
	\subsection{Fundamental Open Questions and Main Results}
	As discussed in Section~\ref{sec:main-approaches}, current methods for addressing heavy-tailed noise predominantly employ either gradient clipping or a combination of gradient normalization and clipping. The ubiquitous utilization of gradient clipping in these approaches raises the following fundamental question: 
	\begin{itemize}
		\item[Q1.] \textit{Is gradient clipping genuinely indispensable for handling heavy-tailed noise? Can gradient normalization alone ensure SGD convergence with heavy-tailed noise?}
	\end{itemize}
	
	This paper investigates the convergence properties of gradient normalization, both with and without variance reduction, in the presence of heavy-tailed noise. Our analysis demonstrates that gradient normalization alone suffices to guarantee SGD convergence under heavy-tailed noise, thereby addressing Q1. However, this result raises a further question: given that either gradient normalization or gradient clipping alone can ensure SGD convergence under heavy-tailed noise, is it necessary to combine these techniques in algorithmic development? Specifically,
	\begin{itemize}
		\item[Q2.] \textit{Does combining gradient normalization with clipping offer additional advantages over using either technique alone?}
	\end{itemize}

	This paper provides the affirmative answer to this question without any mini-batch size requirement. We demonstrate that combining gradient normalization with clipping (NSGDC, Algorithm~\ref{alg-NSGDC}) substantially improves upon the convergence properties of SGD with gradient clipping alone (SGDC, Algorithm~\ref{alg0}) \citep{zhang2020adaptive} under heavy-tailed noise. Our analysis reveals two key advantages. First, NSGDC eliminates the logarithmic factors that affect the convergence rate in SGDC \citep{zhang2020adaptive}, achieving an optimal convergence rate that matches the theoretical lower bound. Second, we establish that NSGDC significantly outperforms SGDC, with the performance gap growing arbitrarily large as the gradient noise $\sigma^2$ approaches zero. Moreover, our analysis proves tight, as it successfully recovers the convergence rate in the deterministic setting where $\sigma^2 = 0$.
We prove that NSGDC achieves convergence rates comparable to those of SGD with gradient normalization alone, while relying on global Lipschitz assumptions. Notably, both methods enjoy convergence guarantees without any requirement on the mini-batch size. This property is particularly appealing from a generalization perspective, as large mini-batches have been shown to negatively affect generalization performance \citep{keskar2017large}.
 These findings together comprehensively address Q2. All these convergence as mentioned above results are summarized in Table~\ref{table-1}. 
		\begin{table}[t]
			\caption{\small A comprehensive comparison between our results and existing convergence rates for SGD under heavy-tailed noise, including \citep{liu2023breaking,cutkosky2021high,nguyen2023improved,zhang2020adaptive}. ``BG'' indicates the assumption of bounded gradient with constant $G>0$ ($G$ is always much larger than $\sigma$), ``GL'' means the global gradient Lipschitz, and    ``IL" is the individual Lipschitz (Assumption \ref{ass1}). Our convergence rate is characterized by  $\frac{1}{T}\sum_{t=1}^T\EE\|\nabla f(\bm{w}^{t})\|$,  the convergence rate in \citep{zhang2020adaptive} is characterized by  $\frac{1}{T}\sum_{t=1}^T\EE\|\nabla f(\bm{w}^{t})\|^2$, while others are characterized by  $\frac{1}{T}\sum_{t=1}^T\|\nabla f(\bm{w}^{t})\|$ with probability $1-\delta$. Our rate significantly outperforms those of \citep{cutkosky2021high,liu2023breaking} when $\sigma \neq 0$. Additionally, for very small values of $\sigma$, our rates show substantial improvement over \citep{liu2023breaking,cutkosky2021high,nguyen2023improved,zhang2020adaptive}, particularly for $1 \leq p < 2$, with the advantage being most pronounced as $p$ approaches 1. {\color{black}Our convergence rates  are established in expectation, leading to improvements over previous high-probability results, which exhibit a gap in the dependence on $\sigma$ even when $p=2$.}}
			\centering
			\begin{tabular}{lclc}
				\toprule 
				\textbf{Algorithms} & \textbf{Papers} &  \textbf{Rate} & \textbf{Assumptions} \\ \midrule
				SGDC                    &		\cite{zhang2020adaptive}                          & $\mathcal{O}\left(\frac{1}{T^{\frac{p-1}{3p-2}}}+\frac{1}{T^{\frac{2p-p^2}{3p-2}}\sigma^{\frac{2p^2}{3p-2}}}\right)$                & GL      \vspace{2mm}  \\
				SGDC                    &\cite{nguyen2023improved}                 & $\mathcal{O}\left(\frac{\sigma\ln\frac{1}{\delta}}{T^{\frac{p-1}{3p-2}}}+\frac{\ln\frac{1}{\delta}}{T^{\frac{p-1}{3p-2}}}\right)$         & GL \vspace{2mm}
                \\
                \rowcolor{limegreen}{NSGD}                   &	Theorem 2                         & $\mathcal{O}\left(\frac{{\color{black}\sigma^{\frac{2p-2}{3p-2}}}}{T^{\frac{p-1}{3p-2}}}+\frac{1}{T^{1/2}}\right)$                & IL \vspace{2mm}\\
                NSGDC                    & 	\cite{cutkosky2021high}                    & $\mathcal{O}\left(\frac{G\ln\frac{ T}{\delta}}{T^{\frac{p-1}{3p-2}}}\right)$                                                     & BG, GL   \vspace{2mm}   \\
				NSGDC                    &	\cite{liu2023breaking}                   & $\mathcal{O}\left(\max\Big\{\frac{\sigma\ln\frac{T}{\delta}}{T^{\frac{p-1}{3p-2}}},\frac{1}{T^{\frac{p-1}{3p-2}}}\Big\}\right)$                                      & GL             \vspace{2mm}     \\ 
			\rowcolor{orchid}{	NSGDC}                    &		Theorem 3                          & $\mathcal{O}\left(\frac{\sigma^{\frac{p}{3p-2}}}{T^{\frac{p-1}{3p-2}}}+\frac{1}{T^{1/2}}\right)$              & GL         \\
				\bottomrule
			\end{tabular}
			\label{table-1}
		\end{table}

	The final open question pertains to momentum acceleration in SGD. While extensive research on accelerated SGD with bounded gradient noise exists, studies on accelerated SGD under heavy-tailed noise remain limited. Given this context, a natural question arises:
	\begin{itemize}
		\item[Q3.] \textit{Can NSGDC achieve accelerated convergence in the presence of heavy-tailed noise?}
	\end{itemize}
	
	This paper introduces an accelerated variant of NSGDC that employs two momentum terms. Under a second-order Lipschitz condition, we prove that this new algorithm converges significantly faster than the standard NSGDC, thus addressing Q3.

\paragraph{Clipping: necessary or beneficial?}
The above questions aim to clarify a fundamental issue in optimization with heavy-tailed noise: is gradient clipping truly necessary for convergence, or merely helpful for acceleration? This work provides a unified view. We do not argue that clipping should always be removed; rather, we identify the regimes in which it is unnecessary, and others where it provably improves convergence rates. Our results span multiple algorithmic variants -- both with and without clipping -- and aim to characterize its theoretical necessity and utility in a principled manner.

	\subsection{Contributions} 
	Our contributions can be summarized as follows.
	\begin{itemize}
		\item We investigate the convergence of gradient normalization in the presence of heavy-tailed noise. Our analysis shows that gradient normalization alone ensures SGD convergence in this setting, addressing Q1.
		
		\item We present a refined convergence analysis for NSGDC and NSGDC-VR, showing that combining gradient normalization with gradient clipping offers additional advantages beyond employing either technique alone, addressing Q2. Previously, the best-known convergence rates for NSGDC and NSGDC-VR from \citep{liu2023breaking} matched those of SGDC \citep{zhang2020adaptive}, failing to justify any theoretical benefits of combining normalization and clipping.
		
		\item We propose an enhanced NSGDC variant incorporating two momentum terms, achieving substantially faster convergence rates than the standard NSGDC algorithm, addressing Q3.

	\end{itemize}

	\subsection{More Related Works}
	\textbf{SGD under Heavy-tailed Noise.}
	In \citep{zhang2020adaptive}, the authors first establish the convergence of SGD with gradient clipping under heavy-tailed noise (Assumption \ref{ass2}). They demonstrate that the convergence rate, in expectation, for $\frac{\sum_{t=1}^T\EE\|\nabla f(\bm{w}^t)\|}{T}$ is $\mathcal{O}(T^{-\frac{p-1}{3p-2}})$ in the nonconvex case. This rate matches the lower bound for nonconvex stochastic optimization, also proven in the same work for cases where the global function $\nabla f(\cdot)$ is Lipschitz.
	\citep{sadiev2023high} investigates the application of stochastic algorithms to solve variational inequalities under heavy-tailed noise, a generalization that extends the applicability of these methods beyond traditional stochastic optimization problems. In \citep{cutkosky2021high}, the authors prove a high-probability convergence rate of $\mathcal{O}\Big(\frac{\ln T}{T^{\frac{p-1}{3p-2}}}\Big)$ when using a combination of gradient normalization and clipping, with the logarithmic factor being eliminated in \citep{liu2023breaking}.
	Further improvements are presented in \citep{puchkin2024breaking}, where the authors assume a specific structure for the heavy-tailed noise and achieve enhanced results. Additionally, paper \citep{vural2022mirror} examines the convergence rate of the Stochastic Mirror Descent algorithm in stochastic convex optimization with heavy-tailed noise, providing lower bounds that emphasize the algorithm's efficiency without the need for gradient clipping.  In the latter work \citep{nguyen2023improved}, the authors offer an improved convergence rate for SGD with clipping under heavy-tailed noise for convex cases.  {\color{black}In \citep{sun2025nonconvex}, the authors prove  that SGD with gradient normalization and momentum   still converges for a class of unbounded variance noise even under delayed updates and weakened smoothness assumptions.}

    \vspace{1mm}
    \noindent \textbf{Nonconvex Variance Reduction.}
Variance reduction techniques have gained significant attention as effective strategies for improving the efficiency of stochastic optimization, particularly in convex finite-sum problems \citep{johnson2013accelerating, shalev2013stochastic, defazio2014saga}. Building on this foundation, researchers have extended these methods to nonconvex smooth optimization. Among the most influential works, \citet{zhou2020stochastic} proposed a nested variance-reduced stochastic gradient algorithm that introduces multi-level reference point construction to control gradient variance more tightly. This design achieves improved complexity guarantees for reaching both first- and second-order stationary points and provides a unified framework for analyzing variance reduction in nonconvex settings.
In parallel, \citet{fang2018spider} introduced the another variance reduction algorithm, which also targets variance reduction for nonconvex finite-sum problems and achieves near-optimal sample complexity. 
Complementing these algorithmic developments, \citet{zhou2019lower} established lower bounds for smooth nonconvex finite-sum optimization, showing that many first-order variance-reduction methods are close to optimal up to logarithmic factors. These results provide both a theoretical benchmark and practical guidance for future algorithm design.
Subsequent works \citep{zhou2018stochastic, cutkosky2019momentum, tran2019hybrid, zhou2020stochasticAI, li2021page} have explored simpler and more robust variants, extending the applicability of variance reduction to broader settings.
More recently, \citet{liu2023breaking} extended variance-reduction techniques to optimization under heavy-tailed noise, addressing the limitations of standard assumptions in stochastic gradient methods.

    \vspace{1mm}
     \noindent\textbf{SGD under Additional Second-Order Smoothness.}
	In stochastic settings, nonconvex accelerated algorithms that rely on both first- and second-order smoothness typically achieve a convergence rate of $\mathcal{O}(T^{-\frac{2}{7}})$ \citep{tripuraneni2018stochastic,allen2018natasha,fang2019sharp,zhou2018finding,sun2023rethinking,cutkosky2020momentum}. 
	When compared to nonconvex variance reduction algorithms, nonconvex accelerated algorithms follow a different path in terms of their smoothness requirements. While variance reduction methods primarily focus on reducing stochastic noise in gradients to improve convergence, nonconvex accelerated algorithms take advantage of higher-order smoothness. Specifically, these algorithms do not demand increased first-order smoothness, such as the Lipschitz continuity of the stochastic gradient. Instead, they rely more heavily on second-order smoothness properties, i.e., the Lipschitz continuity of the Hessian, to ensure acceleration.
	
    \vspace{1mm}
     \noindent\textbf{Two concurrent works.} We note that two concurrent studies independently investigate gradient normalization alone under heavy-tailed noise \citep{hubler2024gradient, liu2024nonconvex}, both of which appeared on arXiv around the same time as our paper\footnote{\citep{hubler2024gradient} was posted on October 17, 2024; \citep{liu2024nonconvex} was posted on December 27, 2024; and the first version of our paper was posted  on October 21, 2024.}.
In \citep{hubler2024gradient}, the authors show that gradient normalization with mini-batches guarantees the convergence of SGD under heavy-tailed noise, achieving the optimal lower bound on sample complexity. They also provide high-probability bounds for the convergence rate.
In   work \citep{liu2024nonconvex}, the authors incorporate momentum into mini-batch gradient normalization for SGD under heavy-tailed noise and establish convergence rates in expectation. {\color{black}Both concurrent works operate under the GL assumption. This paper
treats both regimes: our IL-based results focus on the effect of normalization-only and variance reduction algorithms under heavy-tailed noise, while our GL-based results (Theorem 3) combine normalization with
clipping and compete head-on with \cite{liu2024nonconvex,hubler2024gradient}.}

Our work differs in several aspects from these two studies. First, we do not focus solely on gradient normalization for heavy-tailed noise; instead, our paper also considers several other algorithmic variants. Second, our method imposes no constraint on the mini-batch size,  which naturally leads to a different proof technique.  The minimal requirement on mini-batch size is particularly beneficial for generalization, as large mini-batches have been shown to degrade generalization performance \citep{keskar2017large}. Third, our findings highlight the benefit of jointly applying gradient normalization and clipping.

	\medskip
	
	\noindent\textbf{Notation:}
	This paper uses  $\EE[\cdot]$ to denote the expectation with respect to the underlying probability space. We  use  $\|\cdot\|$ to denote the $L_2$-norm of a vector and  $\|\cdot\|_{\textrm{op}}$ to denote the spectral norm of a matrix. 
	The minimum value of the function $f$  is denoted as $\min f$. Given two sequences $\{a_t\}$ and $\{b_t\}$, we write $a_t=\mathcal{O}(b_t)$ if there exists  a positive constant $0<C<+\infty$ such that
	$a_t \leq C b_t$, and write $a_t=\Theta(b_t)$ if $a_t=\mathcal{O}(b_t)$ and $b_t=\mathcal{O}(a_t)$. Given an event $\mathcal{E}$, we define \( 1_{\mathcal{E}} \)   as the indicator function that takes the value 1 if $\mathcal{E}$ happens and 0 otherwise.

	\section{Assumptions and Technical Novelty}
	\subsection{Other Necessary Assumptions}
	Besides the heavy-tailed noise, we need other basic assumptions utilized for the analysis in our paper. The first assumption pertains to first-order smoothness and unbiased noise, commonly employed within the stochastic community.
	
	\medskip
	
	\noindent \textbf{Assumption ~2}'\emph{
		There exists   constant $L>0$, for any  $\bm{x},\bm{y}\in\RR^d$,
		function $f(\cdot)$ obeys
		\begin{align*}
			\|\nabla f({\bm y})-\nabla  f({\bm x})\|\leq L\|{\bm y}-{\bm x}\|.
		\end{align*}
		We assume stochastic gradient $\bm{g}^t$ satisfies
		\begin{align*}
			\mathbb{E}\bm{g}^t&= \nabla f(\bm{w}^t).
		\end{align*}
		The function value of $f$ satisfies $\inf f>-\infty$.}
	\medskip
	
	\noindent Assumption ~2' is indeed the global Lipschitz, which is quite general as it only requires the gradient of the objective function $f$ to be Lipschitz. In many cases, a stronger assumption can be applied, where the gradient of the objective function $f(\cdot;\xi)$ is Lipschitz (referred to as ``\emph{individual Lipschitzness}" in this paper).
	\begin{assumption}\label{ass1}
		There exists   constant $L>0$, for any  $\bm{x},\bm{y}\in\RR^d$,
		function $f(\cdot;\xi)$ obeys
		\begin{align*}
			\|\nabla f({\bm y};\xi)-\nabla  f({\bm x};\xi)\|\leq L\|{\bm y}-{\bm x}\|.
		\end{align*}
		We assume stochastic gradient $\nabla f(\cdot;\xi)$ satisfies
		\begin{align*}
			\mathbb{E}_{\xi\sim \mathcal{D}} [\nabla f(\bm{w};\xi)] &= \nabla f(\bm{w}),
		\end{align*}
		for any $\bm{w}\in \RR^d$.  
 In addition, we assume that the stochastic gradient is  bounded at the initialization point $w_0$ as
\begin{align}\label{cond}
			B:=\sup_{\xi\sim \mathcal{D}}\{\|\nabla f(\bm{w}^0;\xi)\|\}<+\infty, ~a.s.
		\end{align}
for some $B>0$. The function value of $f(\cdot;\xi)$ satisfies $\inf f(\bm{w};\xi)>-\infty$.
	\end{assumption}
	Compared to Assumption ~2', Assumption \ref{ass1} employs an additional requirement with respect to the boundedness of the initial point $\bm{w}^0$, i.e., \eqref{cond}. We provide justification for the reasonability of this condition. The Lipschitz continuity of $\nabla f(\cdot;\xi)$ indicates
	$$\|\nabla f(\bm{w}^0;\xi)\|^2\leq 2L[ f(\bm{w}^0;\xi)-\min_{\bm{w}} f(\bm{w};\xi)]\leq 2L[ f(\bm{w}^0;\xi)-\inf f(\bm{w};\xi)]. $$
	If   the measurement of the set
	$$\bm{V}:=\{\xi\sim \mathcal{D}\mid \|\nabla f(\bm{w}^0;\xi)\|=+\infty\}$$
	is nonzero. Hence, we can get
	$$+\infty=\int_{\xi\in\bm{V}}\|\nabla f(\bm{w}^0;\xi)\|^2 m(d\xi)\leq \EE \|\nabla f(\bm{w}^0;\xi)\|^2\leq 2L\EE[ f(\bm{w}^0;\xi)-\inf f(\bm{w};\xi)],$$
	indicating that initialization point $\bm{w}^0$ satisfies $f(\bm{w}^0)=\EE[ f(\bm{w}^0;\xi)]=+\infty$.  Hence, the set $\bm{V}$ must have measure zero, and we use condition \eqref{cond}.
	
	In mathematical formulations, the concept of individual Lipschitz continuity closely resembles that of global Lipschitz continuity. However, considering individual Lipschitzness,  we can derive improved lower bounds for stochastic nonconvex optimization problems \citep{carmon2020lower}. The improved lower bound suggests that faster algorithms can be developed for scenarios characterized by this type of smoothness. Indeed, the nonconvex variance reduction method has been specifically designed to take advantage of this smoothness \citep{cutkosky2019momentum,liu2023breaking}.
	
	\subsection{A Concise Explanation of How We Achieve Improved Results}
	This section concisely explains why we can derive improved results for both NSGDC and NSGDC-VR. Previous studies have used complicated high-probability analysis methods, such as \citep{cutkosky2021high, liu2023breaking, nguyen2023improved}. The difficulty arises from the fact that it is challenging to efficiently bound $\EE\|\Ch_{h}(\bm{g}^t)-\nabla f(\bm{w}^t)\|^2$, which is a crucial term in these analyses. 
	Indeed, the gradient normalization technique used in NSGDC and NSGDC-VR provides a useful bound as
	$$\|\nabla f(\bm{w}^t)\|\leq \|\nabla f(\bm{w}^0)\|+L\gamma T.$$
	This inequality allows us to set $h \geq 2(\|\nabla f(\bm{w}^0)\| + L\gamma T)$, ensuring that $\|\nabla f(\bm{w}^t)\| \leq \frac{h}{2}$ for all iterations. 
	By leveraging this bound, we simplify the analysis significantly. According to [Lemma 5, \citep{liu2023breaking}], we can establish the following bounds:
	\begin{align*}
		\EE\|\Ch_{h}(\bm{g}^t)-\EE\Ch_{h}(\bm{g}^t)\|^2 \leq 10\sigma^p h^{2-p}, 
		\|\EE\Ch_{h}(\bm{g}^t)-\nabla f(\bm{w}^t)\|^2 \leq 4\sigma^{2p}h^{-2(p-1)}.
	\end{align*}
	We provide further details on these bounds in Lemma \ref{lemma-c}. Using these bounds, we can then present the improved results.
	
	\medskip
	
	For NSGD and NSGD-VR, the analysis focuses on bounding $\EE\|\bm{g}^t - \nabla f(\bm{w}^t)\|^2$. By leveraging individual Lipschitz smoothness and gradient normalization, we can establish the following inequality
	$$\|\bm{g}^t - \nabla f(\bm{w}^t)\| \leq 2(B + L\gamma T),$$
	which leads to the bound
	$$\EE\|\bm{g}^t - \nabla f(\bm{w}^t)\|^2 \leq [2(B + L\gamma T)]^{2-p} \EE\|\bm{g}^t - \nabla f(\bm{w}^t)\|^p \leq [2(B + L\gamma T)]^{2-p}\sigma^p.$$
	Further details of this bound are provided in Lemma \ref{lemma0}. These bounds significantly simplify the proof process by reducing the complexity of the analysis.
	\section{Gradient Normalization Alone Guarantees Convergence of SGD under Heavy-Tailed Noise}\label{sec-in}
	We present the scheme of NSGD in Algorithm \ref{alg1}. Compared with Algorithm \ref{alg1-c}, Algorithm \ref{alg1} can be regarded as setting $h = +\infty$ in Algorithm \ref{alg1-c}. This adjustment eliminates the clipping hyperparameter in Algorithm \ref{alg1-c}, leading to simpler analysis and more straightforward practical implementations.
	\begin{algorithm}[H]
		\caption{Normalized Stochastic Gradient Descent (NSGD) \citep{cut2020momentum}}\label{alg1}
		\begin{algorithmic}[1]
			\REQUIRE   parameters $\gamma>0$,  $0\leq \theta<1$\\
			\textbf{Initialization}: $\bm{w}^0=\bm{w}^1$, $\bm{m}^0=\bm{0}$\\
			\textbf{for}~$t=1,2,\ldots$ \\
			~~\textbf{step 1}: Sample the data  $\xi^t\sim\mathcal{D}$ and 
			$\bm{m}^t=\theta\bm{m}^{t-1}+(1-\theta)\nabla f(\bm{w}^t;\xi^t)$\\
			~~\textbf{step 2}: $ \bm{w}^{t+1}= \bm{w}^{t}-\gamma \frac{\bm{m}^t}{\|\bm{m}^t\|}$ \\
			\textbf{end for}\\
		\end{algorithmic}
	\end{algorithm}
	Similar to the NSGDC-VR, we can also introduce the variance reduction scheme of NSGD in Algorithm \ref{alg2}, which, like its counterpart, sets the hyperparameter $h$ to $+\infty$ in Algorithm \ref{alg2-c}. This modification effectively simplifies the algorithm while maintaining its variance reduction capabilities.
	\begin{algorithm}[H]
		\caption{Normalized Stochastic Gradient Descent with Variance Reduction (NSGD-VR)}\label{alg2}
		\begin{algorithmic}[1]
			\REQUIRE   parameters $\gamma>0$,  $0\leq \theta<1$\\
			\textbf{Initialization}: $\bm{w}^0=\bm{w}^1$, $\bm{m}^0=\bm{0}$\\
			\textbf{for}~$t=1,2,\ldots$ \\
			~~\textbf{step 1}: Sample the data  $\xi^t\sim\mathcal{D}$ and
			$\bm{m}^t=\theta\bm{m}^{t-1}+\nabla f(\bm{w}^t;\xi^t)-\theta\nabla f(\bm{w}^{t-1};\xi^t)$\\
			~~\textbf{step 2}: $ \bm{w}^{t+1}= \bm{w}^{t}-\gamma \frac{\bm{m}^t}{\|\bm{m}^t\|}$ \\
			\textbf{end for}\\
		\end{algorithmic}
	\end{algorithm}
	
	\subsection{NSGD under Heavy-tailed Noise}
	We present the convergence rate of NSGD under the heavy-tailed noise in the following theorem.
	\begin{theorem}\label{th1}
		Let $(\bm{w}^t)_{t\geq0}$ be generated by Algorithm \ref{alg1}. Suppose Assumptions \ref{ass2}-\ref{ass1} hold.
		If $1-\theta=\min\Big\{\frac{{\color{black}\max\{(L\Delta)^{\frac{1}{2}}, 1\}}}{\sigma^{\frac{4p-4}{3p-2}}T^{\frac{p}{3p-2}}},1\Big\}$, $\gamma={\color{black}\sqrt{\frac{\Delta}{L}}}\frac{\sqrt{1-\theta}}{\sqrt{T}}$ {\color{black}with $\Delta:=f(\bm{w}^0)-\min f$},
		then it holds that
        {\color{black}
	\begin{align*}
		\begin{aligned}
			\frac{1}{T}\sum_{t=1}^T\EE\|\nabla f(\bm{w}^{t})\|&\leq \frac{{\color{black}4(L\Delta)^{\frac{1}{4}}}\sigma^{\frac{2p-2}{3p-2}}}{T^{\frac{p-1}{3p-2}}}+\frac{2\|\nabla f(\bm{w}^0)\|\sigma^{\frac{4p-4}{3p-2}}}{T^{\frac{2p-2}{3p-2}}}\\
&+\frac{4{\color{black}(L\Delta)^{\frac{6-3p}{8}}}\Big(\frac{B}{L}+1\Big)^{\frac{2-p}{2}}\sigma^{\frac{5p^2-12p+8}{6p-4}}}{T^{\frac{(p-2)^2}{2(3p-2)}+\frac{p-1}{3p-2}}}+\frac{4{\color{black}(L\Delta)^{\frac{2-p}{4}}}\Big(\frac{B}{L}+1\Big)^{\frac{2-p}{2}}\sigma^{\frac{5p^2-12p+8}{6p-4}}}{T^{\frac{(p-2)^2}{2(3p-2)}+\frac{p-1}{3p-2}}}\\
			&+\frac{4{\color{black}(L\Delta)^{\frac{2-p}{4}}}\Big(\frac{B}{L}+1\Big)^{\frac{2-p}{2}}\sigma^{p/8}}{T^{\frac{1}{4}}}+\frac{{\color{black}4\sqrt{L\Delta}}}{T^{\frac{1}{2}}}+\frac{2\|\nabla f(\bm{w}^0)\|}{T}.
		\end{aligned}
	\end{align*}
        }
	\end{theorem}
When $\sigma = 0$, the convergence rate presented above matches that of GD, whereas when $p = 2$, it reduces to the convergence rate of standard SGD as $\mathcal{O}\Big(\frac{\sqrt{\sigma}}{T^{\frac{1}{4}}}\Big)$.
When $\sigma>0$ is large, the right-hand side of the convergence bound can further be simplified as
\[
\mathcal{O}\left(\frac{{\color{black}(L\Delta)^{\frac{1}{4}}\sigma^{\frac{2p-2}{3p-2}}}}{T^{\frac{p-1}{3p-2}}} +\frac{\sigma^{\frac{p}{8}}}{T^{\frac{1}{4}}}+ \frac{1}{T^{1/2}}\right)=\mathcal{O}\left(\frac{ {\color{black}(L\Delta)^{\frac{1}{4}}\sigma^{\frac{2p-2}{3p-2}}}}{T^{\frac{p-1}{3p-2}}}  \right).
\]
When $\sigma = 0$, the convergence rate presented above matches that of GD, whereas when $p = 2$, it reduces to the convergence rate of standard SGD.

	In Theorem \ref{th1}, we show that gradient clipping is unnecessary to guarantee convergence for heavy-tailed noise, which contrasts with the approach in \citep{cutkosky2021high, liu2023breaking} where gradient clipping is used alongside gradient normalization. Removing the gradient clipping eliminates the need to tune the clipping hyperparameter, simplifying the algorithm. The convergence rate we achieve is as fast as that of NSGDC, as demonstrated in Theorem \ref{th1-c} presented in the next section. However, it is essential to note that Theorem \ref{th1} relies on a stronger Lipschitz assumption, specifically individual Lipschitzness rather than global Lipschitzness. This assumption is crucial because it allows us to bound $\|\nabla f(\bm{w}^t;\xi^t)\|$ without relying on gradient clipping in NSGD.  Although individual Lipschitzness is considered a stronger condition than global Lipschitzness, it is highly relevant and natural in the context of learning. This is because the smoothness of the expected objective function typically originates from the loss function $f(\cdot;\xi)$ evaluated at each individual sample.  

 Indeed, the concurrent works on NSGD with heavy-tailed noise \citep{liu2024nonconvex, hubler2024gradient} both establish the convergence of   NSGD  under heavy-tailed noise and global Lipschitz continuity, achieving the same optimal sampling complexity as our method. Notably, \citet{hubler2024gradient} presented a stronger result by providing high-probability convergence guarantees, in contrast to our expectation-based analysis. However,  \citet{hubler2024gradient}  impose specific requirements on the mini-batch size used in the algorithm. In contrast, our convergence guarantees do not rely on any mini-batch assumption. This distinction is particularly relevant in practice, as large mini-batch sizes are known to often degrade generalization performance. Compared with \citep{liu2024nonconvex}, our proven convergence rate demonstrates improved dependence on the noise level, particularly in terms of the power of $\sigma$, especially when   $\sigma$ is large and $p$ is close to 1. Specifically, our result achieves a rate of $\mathcal{O}\left(\frac{ {\color{black}\sigma^{\frac{2p-2}{3p-2}}}}{T^{\frac{p-1}{3p-2}}}\right)$, whereas the convergence rate reported in \citep{liu2024nonconvex} is  $\mathcal{O}\left(\frac{ \sigma^{\frac{p}{3p-2}}}{T^{\frac{p-1}{3p-2}}}\right)$.


	\subsection{NSGD-VR under Heavy-tailed Noise}
	We present the convergence for the variance reduction scheme of NSGD under the heavy-tailed noise.
	\begin{theorem}\label{th2}
		Let $(\bm{w}^t)_{t\geq0}$ be generated by Algorithm \ref{alg2}. Suppose Assumptions \ref{ass2}-\ref{ass1} hold.
		If 	$1-\theta=\min\Big\{\frac{1}{\sigma^{\frac{p}{2p-1}}T^{\frac{p}{2p-1}}},1\Big\}$, $\gamma=\frac{\sqrt[4]{1-\theta}}{L\sqrt{T}}$,
		then it holds that
		\begin{align*}
			\begin{aligned}
				&\frac{1}{T}\sum_{t=1}^T\EE\|\nabla f(\bm{w}^{t})\|
				 	\leq  \frac{2\|\nabla f(\bm{w}^0)\|\sigma^{\frac{p}{2p-1}}+4\Big(\frac{B}{L}+1\Big)^{\frac{2 -p}{2}}\sigma^{\frac{p}{2p-1}}+4\Big(\frac{B}{L}+1\Big)^{\frac{2 -p}{2}}\sigma^{\frac{p}{4(2p-1)}}}{T^{\frac{p-1}{2p-1}}}\\
			&+\frac{[L(f(\bm{w}^{0})-\min f)+5]\sigma^{\frac{p}{4(2p-1)}}}{T^{\frac{p-1}{2p-1}}}+\frac{L(f(\bm{w}^{0})-\min f)+4\Big(\frac{B}{L}+1\Big)^{\frac{2 -p}{2}}+5}{ \sqrt{T}}+\frac{2\|\nabla f(\bm{w}^0)\|}{T}.
			\end{aligned}
		\end{align*}
		
	\end{theorem}
When $\sigma>0$ is large, we can get a simplified rate for NSGD-VR under heavy-tailed noise as
$$\mathcal{O}\left(\frac{ \sigma^{\frac{p}{2p-1}}}{T^{\frac{p-1}{2p-1}}} + \frac{1}{T^{1/2}}\right).$$
 When $p=2$, i.e., in the traditional bounded variance case, the convergence rate of NSGD-VR for $\frac{1}{T}\sum_{t=1}^T \mathbb{E}\|\nabla f(\bm{w}^{t})\|$ is $\mathcal{O}\left(\frac{1}{T^\frac{1}{3}}\right)$ as $\sigma\neq 0$, which matches the speed of STORM \citep{cutkosky2019momentum} but without requiring the bounded gradient assumption. According to \citep{arjevani2023lower}, the lower bound for nonconvex first-order stochastic methods in individually smooth optimization is also $\mathcal{O}\left(\frac{1}{T^\frac{1}{3}}\right)$. Hence, NSGD-VR achieves the lower bound when $p=2$.  
The convergence rate of NSGD-VR is $\mathcal{O}\left(\frac{\sigma^{\frac{p}{2p-1}}}{T^\frac{p-1}{2p-1}}\right)$, which is better than that of \citep{liu2023breaking} as $\mathcal{O}\left(\frac{\sigma\ln T}{T^\frac{p-1}{2p-1}}\right)$. Our results also demonstrate that gradient clipping is unnecessary for heavy-tailed noise to achieve nonconvex variance reduction. Furthermore, compared to \citep{liu2023breaking}, our findings fix the case when $\sigma = 0$, which is not addressed with clipping and normalization. When $\sigma = 0$, the rate is $\mathcal{O}\left(\frac{1}{\sqrt{T}}\right)$, matching the speed of gradient descent for nonconvex optimization. The comparison is described by Table \ref{table-2}.
	\begin{table}[t!]
		\caption{\small A  comparison between Theorem \ref{th2} and existing convergence rates  \citep{liu2023breaking}. Our convergence rate is characterized by $\frac{1}{T}\sum_{t=1}^T\EE\|\nabla f(\bm{w}^{t})\|$, whereas \citep{liu2023breaking} adopts $\frac{1}{T}\sum_{t=1}^T\|\nabla f(\bm{w}^{t})\|$ with a confidence level of at least $1-\delta$.}
		\centering
		\begin{tabular}{lcl}
			\toprule 
			\textbf{Paper} &  \textbf{Rate under $\sigma\neq 0$} & \textbf{Rate under $\sigma= 0$}  \\ \midrule
			\cite{liu2023breaking}                   & $\mathcal{O}\left(\frac{\sigma\ln T}{T^{\frac{p-1}{3p-2}}}\right)$                     & N/A                                \\ 
			Our result                           & $\mathcal{O}\left(\frac{\sigma^{\frac{p}{2p-1}}}{T^{\frac{p-1}{3p-2}}}\right)$               & $\mathcal{O}\left(\frac{1}{T^{1/2}}\right)$       \\
			\bottomrule
		\end{tabular}
		\label{table-2}
	\end{table}

	Indeed, as shown in Theorem \ref{th2}, the convergence rate of NSGD-VR is in the same order as that of NSGDC-VR, as established in Theorem \ref{th2-c} presented in the next section. Both algorithms achieve this rate under the same assumptions, indicating that the clipping procedure is entirely redundant for variance reduction under heavy-tailed noise.
	
	\section{Improved Results of Gradient Normalization with Clipping under Heavy-tailed Noise}\label{sec-improved}
	This section presents improved convergence rates for the gradient normalization with clipping methods. Building on the same assumptions in the most recent paper \citep{liu2023breaking}, we demonstrate that our approaches achieve superior convergence rates.
	\subsection{Improved Convergence  Rate of NSGDC}
	The improved convergence rate of NSGDC is presented by Theorem \ref{th1-c}.
	\begin{theorem}\label{th1-c}
		Let $(\bm{w}^t)_{t\geq0}$ be generated by Algorithm \ref{alg1-c}. Suppose Assumptions \ref{ass2}-2' hold.
		If $1-\theta=\min\Big\{\frac{{\color{black}\max\big\{(L\Delta)^{\frac{p}{3p-2}}, 1\big\}}}{\max\big\{\sigma^{\frac{4p-4}{3p-2}},\sigma^{\frac{2p}{3p-2}}\big\}T^{\frac{p}{3p-2}}},1\Big\}$, $\gamma={\color{black}(1_{\{L\Delta\geq 1\}}\sqrt{\frac{\Delta}{L}}+1_{\{L\Delta< 1\}}\frac{1}{L})}\frac{\sqrt{1-\theta}}{\sqrt{T}}$ {\color{black}with $\Delta:=f(\bm{w}^{0})-\min f$}, $h=2(\gamma LT^{\frac{2p}{3p-2}}\cdot\max\{1,{\color{black}\sigma^{\frac{3p}{3p-2}}\cdot1_{\{\sigma^2T\geq  L\Delta\}}}\}+\|\nabla f(\bm{w}^0)\|)$,
		then it holds that
{\color{black}
     		\begin{align*}
			&\frac{1}{T}\sum_{t=1}^T\EE\|\nabla f(\bm{w}^{t})\|
\leq\frac{4(L\Delta)^{\frac{p-1}{3p-2}} \sigma^{\frac{p}{3p-2}}+23\sigma^{\frac{p}{3p-2}}+3\sigma^{\frac{2p-2}{3p-2}}+16\sigma^{\frac{p-1}{3p-2}}+16\sigma^{\frac{p}{8}}}{T^{\frac{p-1}{3p-2}}}\\
&\qquad +\frac{4 (L\Delta)^{\frac{p-1}{3p-2}}\sigma^{\frac{2p-2}{3p-2}}+16{\color{black} (L\Delta)^{\frac{p}{3p-2}}}\sigma^{\frac{p-1}{3p-2}}+16(L\Delta)^{\frac{6p-2p^2-2}{6p-4}}\sigma^{\frac{p^2}{6p-4}}+16(L\Delta)^{\frac{1}{4}}\sigma^{\frac{p}{8}}}{T^{\frac{p-1}{3p-2}}}\\
			&\qquad +\frac{16[(L\Delta)^{\frac{p}{6p-4}}+1]\cdot\|\nabla f(\bm{w}^{0})\|^{\frac{2 -p}{2}}\sigma^{\frac{3p^2-6p+4}{2(3p-2)}}}{T^{\frac{p}{6p-4}}}+\frac{2\|\nabla f(\bm{w}^0)\|\sigma^{\frac{4p-4}{3p-2}}+2\|\nabla f(\bm{w}^0)\|\sigma^{\frac{2p}{3p-2}}}{T^{\frac{2p-2}{3p-2}}}\\
&\qquad +\frac{4\sqrt{L\Delta}+16\|\nabla f(\bm{w}^0)\|^{\frac{2 -p}{2}}+7}{T^{\frac{1}{2}}}+\frac{2\|\nabla f(\bm{w}^0)\|}{T}.
		\end{align*}}
	\end{theorem}
When $\sigma = 0$, the convergence rate described above aligns with that of  GD. In contrast, when $p = 2$, it simplifies to the convergence rate of standard  SGD as $\mathcal{O}\Big(\frac{\sqrt{\sigma}}{T^{\frac{1}{4}}}\Big)$.  As $\sigma$ is large, we can get a simplified rate for NSGDC under the heavy-tailed noise  as
\[
\mathcal{O}\left(\frac{{\color{black}(L\Delta)^{\frac{p-1}{3p-2}}}\sigma^{\frac{p}{3p-2}}}{T^{\frac{p-1}{3p-2}}} +\frac{\sigma^{\frac{2p}{3p-2}}}{T^{\frac{2p-2}{3p-2}}}+ \frac{1}{T^{1/2}}\right)=\mathcal{O}\left(\frac{ {\color{black}(L\Delta)^{\frac{p-1}{3p-2}}}\sigma^{\frac{p}{3p-2}}}{T^{\frac{p-1}{3p-2}}}  \right).
\]
{\color{black}The presented bound exhibits the same dependence on 
$(L\Delta)^{\frac{p-1}{3p-2}}$
 as established in \citep{liu2024nonconvex}.
 We also revise Theorem 1 to align the dependence on $L\Delta$.
 }

	We present a discussion of our proven results compared to several closely related works \citep{cutkosky2021high, liu2023breaking, zhang2020adaptive,nguyen2023improved}.


First, we establish convergence rates under either weaker or comparable assumptions. Specifically, when $\sigma \neq 0$, our dominant rate achieves $\mathcal{O}\left(\frac{1}{T^{\frac{p-1}{3p-2}}}\right)$ in expectation, without relying on the uniformly bounded gradient assumption. 
In comparison, \citep{cutkosky2021high} and \citep{liu2023breaking} obtain high-probability convergence rates of the form $\mathcal{O}\left(\frac{\ln\frac{T}{\delta}}{T^{\frac{p-1}{3p-2}}}\right)$ with probability at least $1-\delta$ for a small $\delta>0$. Similarly, \citep{nguyen2023improved} establishes high-probability guarantees under heavy-tailed noise, incurring additional logarithmic dependence on $\delta$.
Note that high-probability bounds are generally stronger than expectation bounds, and the additional logarithmic factors are standard consequences of union bound arguments across iterations. In contrast, our result is characterized in expectation, and thus naturally avoids logarithmic terms such as $\ln T$, providing a clean dependence on $T$ without affecting the probabilistic level of guarantee.
Accordingly, our expectation-based convergence result and existing high-probability results address different aspects of algorithmic behavior, and should be interpreted within their respective probabilistic frameworks.
	
	Second, compared to all four works mentioned above \citep{liu2023breaking,cutkosky2021high,nguyen2023improved,zhang2020adaptive}, our results explicitly characterize the subordinate rate. Note that when $\sigma = 0$ (i.e., the deterministic case for any $p \in (1,2]$), the rate is $\mathcal{O}\left(\frac{1}{\sqrt{T}}\right)$ for $\frac{1}{T}\sum_{t=1}^T\|\nabla f(\bm{w}^{t})\|$, matching the convergence rate of gradient descent. In contrast, none of the previous works recover the deterministic results when $\sigma = 0$. We list the detailed comparison in Table \ref{table-1}.  The result also explains the necessity of using gradient normalization in clipping SGD when $\sigma$ is tiny.

{\color{black}Third, when the noise level  $\sigma$ is large, we derive a convergence rate of $\mathcal{O}\left(\frac{ \sigma^{\frac{p}{3p-2}}}{T^{\frac{p-1}{3p-2}}}\right)$, which   improves upon the $\widetilde{\mathcal{O}}\left(\frac{\sigma}{T^{\frac{p-1}{3p-2}}}\right)$ result in high-probability form shown in \citep{liu2023breaking}, since the exponent 
$
\frac{p}{3p - 2}<1   
$ for all  $p \in (1, 2)$. This demonstrates a substantial gain in noise robustness offered by our approach. Indeed, the rate matches the lower bound under heavy-tailed noise with respect to both $\sigma$ and $1/T$ \citep{liu2024nonconvex}.}

Compared with concurrent work on  NSGD \citep{hubler2024gradient, liu2024nonconvex}, our analysis of NSGDC underscores the benefits of combining gradient clipping with normalization. Notably, relative to \citet{hubler2024gradient}, NSGDC eliminates the need for the mini-batch assumption and achieves improved dependence on the noise level $\sigma$ in terms of gradient computation cost. Furthermore, while \citet{liu2024nonconvex} establish a convergence rate of $\mathcal{O}\left(\frac{\sigma^{\frac{p}{3p-2}}}{T^{\frac{p-1}{3p-2}}} + \frac{\sigma}{T^{\frac{p-1}{2p-1}}}\right)$ for NSGD, our results show that NSGDC achieves $\mathcal{O}\left(\frac{\sigma^{\frac{p}{3p-2}}}{T^{\frac{p-1}{3p-2}}} + \frac{\sigma^{\frac{2p}{3p-2}}}{T^{\frac{2p-2}{3p-2}}} \right)$, offering a more favorable subordinate term.

	\subsection{Improved Convergence  Rate of NSGDC-VR}
	We also present the improved convergence result for NSGDC-VR in Theorem \ref{th2-c}.
	\begin{theorem}\label{th2-c}
		Let $(\bm{w}^t)_{t\geq0}$ be generated by Algorithm \ref{alg2-c}. Suppose Assumptions \ref{ass2}-\ref{ass1} hold.
		If $1-\theta=\min\Big\{\frac{1}{\max\big\{\sigma^{\frac{p}{2p-1}},\sigma^{\frac{2p-2}{2p-1}}\big\}T^{\frac{p}{2p-1}}},1\Big\}$, $\gamma=\frac{\sqrt[4]{1-\theta}}{L\sqrt{T}}$, $h=2(\gamma LT^{\frac{5p+2}{8p-4}}\cdot\max\{1,\sigma^{\frac{9p}{8p-4}}\}+\|\nabla f(\bm{w}^0)\|)$,
		then it holds that
		\begin{align*}
			&\frac{1}{T}\sum_{t=1}^T\EE\|\nabla f(\bm{w}^{t})\|
			\leq \frac{20\sigma^{\frac{p}{2p-1}}+2\|\nabla f(\bm{w}^0)\|(\sigma^{\frac{p}{2p-1}}+\sigma^{\frac{2p-2}{2p-1}})+\sigma^{\frac{p}{16}}}{T^{\frac{p-1}{2p-1}}}\\
&\quad+\frac{[L(f(\bm{w}^{0})-\min f)+5](\sigma^{\frac{p-1}{4p-2}}+\sigma^{\frac{p}{8p-4}})}{ T^{\frac{3p-2}{8p-4}}}+\frac{16\|\nabla f(\bm{w}^{0})\|^{\frac{2-p}{2}}(\sigma^{\frac{p^2-p}{2p-1}}+\sigma^{\frac{2p^2-5p+2}{4p-2}})}{T^{\frac{p}{4p-2}}}\\
			&\quad+\frac{16\sigma^{\frac{5p^2-9p+6}{8p-4}}}{T^{\frac{5p-1}{8p-4}}}+\frac{L(f(\bm{w}^{0})-\min f)+16\|\nabla  f(\bm{w}^{0})\|^{\frac{2-p}{2}}+9}{ T^{\frac{1}{2}}}+\frac{2\|\nabla f(\bm{w}^0)\|}{T}.
		\end{align*}	
	\end{theorem}
When $\sigma = 0$, the convergence rate proved above corresponds to that of GD. On the other hand , if $p = 2$, it simplifies to the convergence rate of standard SGD, expressed as $\mathcal{O}\Big(\frac{\sqrt{\sigma}}{T^{\frac{1}{4}}}\Big)$. When $\sigma$ becomes large,   we derive a simplified rate for NSGDC-VR as
    $$\mathcal{O}\left(\frac{\sigma^{\frac{p}{2p-1}}}{T^{\frac{p-1}{2p-1}}} + \frac{1}{T^{1/2}}\right).$$
{\color{black}The rate in Theorem \ref{th2-c} is on par with that of Theorem \ref{th2}. We stress the ``improved result’’ refers to the comparison with \citep{liu2023breaking}. More precisely}, unlike previous results on variance reduction algorithms under heavy-tailed noise \citep{liu2023breaking}, our work successfully characterizes the subordinate term, which was not addressed in earlier studies.
 When $\sigma \neq 0$, our method achieves a convergence rate as fast as that established in \citep{liu2023breaking} as $\mathcal{O}\Big(\frac{1}{T^{\frac{p-1}{2p-1}}}\Big)$. Moreover, when $\sigma = 0$, our result matches the convergence rate of gradient descent in the deterministic case, a scenario not covered in prior research. When $\sigma$ is large, our rate $\mathcal{O}\left(\frac{\sigma^{\frac{p}{2p-1}}}{T^{\frac{p-1}{2p-1}}}\right)$ is better than that in \citep{liu2023breaking}, which is $\mathcal{O}\left(\frac{\sigma}{T^{\frac{p-1}{2p-1}}}\right)$, since $\frac{p}{2p-1} \leq 1$ as $p\in(1,2]$.

	\section{Accelerations with Second-order Smoothness under Heavy-tailed Noise}\label{sec-2}
	This section explores how second-order smoothness enhances the convergence rate of SGD under heavy-tailed noise. We present an approach inspired by the accelerated algorithm introduced in \citep{cutkosky2020momentum}.
	\begin{algorithm}[H]
		\caption{Accelerated Normalized Stochastic Gradient Descent with Clipping under Second-Order Smoothness (A-NSGDC)}\label{alg3-c}
		\begin{algorithmic}[1]
			\REQUIRE   parameters $\gamma>0$,  $0\leq \theta<1$, $\zeta\geq 0$\\
			\textbf{Initialization}: $\bm{w}^0=\bm{w}^1$, $\bm{m}^0=\bm{0}$\\
			\textbf{for}~$t=1,2,\ldots$ \\
			~~\textbf{step 1} $\bm{v}^t=\bm{w}^{t}+\zeta(\bm{w}^{t}-\bm{w}^{t-1})$\\
			~~\textbf{step 2}: Sample unbiased stochastic gradient $\EE\bm{g}^t=\nabla f(\bm{v}^t)$\\
			~~\textbf{step 3}: Updates $\bm{m}^t=\theta\bm{m}^{t-1}+(1-\theta)\Ch_{h}(\bm{g}^t)$ \\
			~~\textbf{step 4}: $ \bm{w}^{t+1}= \bm{w}^{t}-\gamma \frac{\bm{m}^t}{\|\bm{m}^t\|}$ \\
			\textbf{end for}\\
		\end{algorithmic}
	\end{algorithm}
	The acceleration of A-NSGDC is proved under the second-order Lipschitz continuity.
	\begin{assumption}\label{ass3}
		There exists $H>0$ such that the Hessian matrix of function $f$ satisfies $\|\nabla^2 f(\bm{x})-\nabla^2 f(\bm{y})\|_{\emph{op}}\leq H\|\bm{x}-\bm{y}\|$ for $\bm{x},\bm{y}\in\RR^d$.
	\end{assumption}
	In the stochastic setting, the second-order Lipschitz continuity is commonly employed for analyzing accelerated nonconvex optimization, as demonstrated in \citep{tripuraneni2018stochastic, allen2018natasha, fang2019sharp, zhou2018finding, sun2023momentum, cutkosky2020momentum, sun2023rethinking}.
	
	In the following, we present the accelerated convergence rate for A-NSGDC under second-order smoothness  with heavy-tailed noise.
	\begin{theorem}\label{th3-c}
		Let $(\bm{w}^t)_{t\geq0}$ be generated by Algorithm \ref{alg3-c}. Suppose Assumptions \ref{ass2}, 2',  and \ref{ass3} hold.
		If 	$1-\theta=\min\Big\{\frac{1}{\max\big\{\sigma^\frac{6}{7},\sigma^\frac{3}{7}\big\} T^{\frac{2p}{4p-1}}},1\Big\}$, $\gamma=\frac{(1-\theta)^{\frac{2}{3}}}{LT^{\frac{2p+3}{12p-3}}}(\frac{1_{\theta=0}}{T^{\frac{8p-9}{24p-6}}}+1_{\theta\neq 0})$,  $h=2(\gamma LT^{\frac{2p+3}{4p-1}}\cdot\max\{1,\sigma^{\frac{13p-10}{7p-7}}\}+\|\nabla f(\bm{w}^0)\|)$, $\zeta=\frac{\theta}{1-\theta}$,
		then it holds that
		\begin{align*}
			 &\frac{1}{T}\sum_{t=1}^T\EE\|\nabla f(\bm{w}^{t})\| 
			\leq \frac{L(\sigma^{\frac{4}{7}}+\sigma^{\frac{2}{7}})(f(\bm{w}^{1})-\min f)+20(\sigma^{\frac{4}{7}}+\sigma^{\frac{2}{7}})+16\sigma^{\frac{p}{16}}}{T^{\frac{2p-2}{4p-1}}}\\
&\quad+\frac{16\|\nabla f(\bm{w}^{0})\|^{\frac{2 -p}{2}}(\sigma^{\frac{4}{7}}+\sigma^{\frac{2}{7}})}{T^{\frac{p}{4p-1}}}+\frac{2\|\nabla f(\bm{w}^0)\|(\sigma^{\frac{6}{7}}+\sigma^{\frac{3}{7}})}{T^{\frac{2p-1}{4p-1}}} +\frac{H(\sigma^{\frac{4}{7}}+\sigma^{\frac{2}{7}})}{L^2T^{\frac{2}{4p-1}}}+\frac{\sigma^{\frac{4}{7}}+\sigma^{\frac{1}{28}}}{2T^{\frac{2p+1}{4p-1}}} \\
			&\quad+\frac{L(f(\bm{w}^{1})-\min f)+5+16\|\nabla f(\bm{w}^{0})\|^{\frac{2 -p}{2}}}{T^{1/2}}+\frac{2\|\nabla f(\bm{w}^0)\|}{T}.
		\end{align*}
	\end{theorem}
In the case where $\sigma = 0$, the convergence rate established above reduces to that of gradient descent (GD). Conversely, when $p=2$, it recovers the standard SGD rate, given by $\mathcal{O}\left(\frac{\sqrt{\sigma}}{T^{1/4}}\right)$. Furthermore, as $\sigma$ increases significantly,
 we can further simplify the rate as
    $$\mathcal{O}\left(\frac{\sigma^{\frac{4}{7}}}{T^{\frac{2p-2}{4p-1}}} + \frac{1}{T^{1/2}}\right).$$
To our knowledge, Theorem \ref{th3-c} provides the first convergence result for an accelerated nonconvex first-order method in the presence of heavy-tailed noise under second-order Lipschitzness. The theorem is proven under a broad smoothness assumption, specifically that the function $f$ is Lipschitz continuous. When $\sigma \neq 0$, Theorem \ref{th3-c} demonstrates a convergence rate of $\mathcal{O}\Big(\frac{1}{T^{\frac{2p-2}{4p-1}}}\Big)$, which is faster than the rate of NSGDC. In particular, when $p=2$, the rate is $\mathcal{O}\Big(\frac{1}{T^{\frac{2}{7}}}\Big)$, aligning with results from \citep{carmon2018accelerated, agarwal2017finding, tripuraneni2018stochastic, allen2018natasha, fang2019sharp, zhou2018finding, sun2023momentum, cutkosky2020momentum, sun2023rethinking}. Furthermore, in the deterministic case where $\sigma = 0$, the convergence rate improves to $\mathcal{O}\Big(\frac{1}{T^{\frac{1}{2}}}\Big)$, matching that of gradient descent.
	
	By setting $ h = +\infty$  in A-NSGDC, we obtain the accelerated version of NSGD under second-order Lipschitz continuity. The same convergence rate as A-NSGDC can be derived following a similar proof strategy (see details in Appendix \ref{NSGD2}). However, we do not further present the result of the accelerated NSGD under second-order Lipschitzness in the main text because NSGD is proved under individual Lipschitz continuity, where variance reduction techniques are more effective. The convergence rate of NSGD-VR is $ \mathcal{O}\left(\frac{1}{T^{\frac{p-1}{2p-1}}}\right)$, which is faster than the $\mathcal{O}\left(\frac{1}{T^{\frac{2p-2}{4p-1}}}\right)$ rate achieved under second-order Lipschitzness. Therefore, applying the accelerated scheme is unnecessary, even in the presence of second-order Lipschitz continuity.

    {\color{black}The structure of our proof for Theorem~\ref{th3-c} is related to that of \citep{cutkosky2021high}, but analyzing the algorithm in expectation provides a complementary perspective. 
In particular, the expectation-based framework allows for a simpler analysis and, more importantly, makes the role of the hyperparameters clearer. This transparency means that hyperparameter choices in the accelerated scheme are more flexible and interpretable, as opposed to high-probability analysis in~\citep{cutkosky2021high}, where the admissible parameter range is quite constrained and technically involved. Therefore, Theorem~\ref{th3-c} should be viewed as complementing the high-probability result of~\citep{cutkosky2021high}: our main contribution lies in demystifying parameter tuning and providing a simplified framework, whereas~\citep{cutkosky2021high} provides the high-probability guarantees.}
	
	A natural question that arises is whether we can integrate the accelerated scheme with variance reduction techniques to further enhance the performance of algorithms under the assumption of second-order smoothness. This combination would ideally allow us to achieve faster convergence. Unfortunately, despite our efforts, we have not yet succeeded in developing such a combined scheme. The main technical challenge stems from a fundamental incompatibility between the accelerated scheme and the variance reduction process. The accelerated scheme approximates the stochastic gradient using a second-order Taylor expansion, which inherently alters the structure of the stochastic gradient. This alteration disrupts the conditions required for effective variance reduction, making it difficult to harmonize these two approaches within a single framework.
	
	\section{Concluding Remarks}
	This work presents significant theoretical advances for NSGDC and NSGDC-VR in addressing nonconvex optimization problems under heavy-tailed noise. Beyond proving improved convergence results, we showed that gradient normalization alone can ensure convergence under heavy-tailed noise without clipping, broadening the applicability of normalization techniques. By leveraging second-order smoothness, we also developed accelerated algorithms that enhance performance in nonconvex settings. These contributions provide new directions for optimizing stochastic algorithms and reinforce the utility of gradient normalization in complicated, noise-affected environments.
	
	Our results can be extended to a broader class of gradient normalizations as defined by \citep{sun2023rethinking}, outlined below.
	
	\begin{definition}
		We define $\Gc$ as a general normalization if, for any $\bm{x} \in \RR^d$, it satisfies
		$$\langle \bm{x}, \Gc(\bm{x}) \rangle \geq l \|\bm{x}\|_{\diamond}, ~\|\Gc(\bm{x})\| \leq U,$$
		where $l, U > 0$, and $\|\cdot\|_{\diamond}$ denotes a specific norm on $\RR^d$.
	\end{definition}
	From this definition, we observe that the sign operator and normalization are particular cases of general normalization. The results established in this paper can thus be directly extended to this broader setting by substituting $\frac{\bm{m}^t}{\|\bm{m}^t\|}$ with $\Gc(\bm{m}^t)$ in all relevant algorithms. The proofs remain nearly identical to that presented in the appendix.

\section*{Acknowledgements}
The authors thank the Editor and the anonymous reviewers for their constructive suggestions. Tao Sun is supported in part by the National Natural Science Foundation of China (Grant Nos. 62376278, 62522610), and NUDT Foundational Research Funding (JS25-02), and by the Young Elite Scientists Sponsorship Program of CAST (No. 2022QNRC001).

	\appendix
	
	\section{Technical Lemmas}
	\begin{lemma}\label{lemma-c}
		Let $(\bm{w}^t)_{t\geq 1}$ be generated by Algorithms \ref{alg1-c}, \ref{alg2-c} and \ref{alg3-c}, if Assumptions \ref{ass2}  and 2' hold and $h\geq 2(\|\nabla f(\bm{w}^{0})\|+L\gamma T)$, we have
		\begin{align}
			\begin{aligned}
				\EE\|\Ch_{h}(\bm{g}^t)-\EE\Ch_{h}(\bm{g}^t)\|^2&\leq 10h^{2-p}\sigma^{p}, \\
				\|\EE\Ch_{h}(\bm{g}^t)-\nabla f(\bm{w}^t)\|&\leq 2\sigma^{p}h^{-(p-1)}.
			\end{aligned}
		\end{align}
	\end{lemma}

	\begin{lemma}\label{lemma0}
		Let $(\bm{w}^t)_{t\geq 1}$ be generated by Algorithms \ref{alg1}, \ref{alg2} and \ref{alg3}, if Assumptions \ref{ass2} and \ref{ass1}  hold, it follows that
		$$\EE_{\xi^t\sim\mathcal{D}}\|\nabla f(\bm{w}^t;\xi^t)-\nabla f(\bm{w}^t)\|^2\leq 4(B+L\gamma T)^{2-p}\sigma^p.$$
	\end{lemma}

	\begin{lemma}\label{lemma1}\citep{cutkosky2020momentum,zhao2021convergence}
		Let $\bm{w}^{\dag},\bm{m}\in\RR^d$ be arbitrary vectors,  and
		\begin{equation}\label{temp-sch-sign}
			\bm{w}^{\ddag}= \bm{w}^{\dag}-\gamma \frac{\bm{m}}{\|\bm{m}\|},
		\end{equation}
		and $\bm{\epsilon}:=\bm{m}-\nabla f(\bm{w}^{\dag})$. When the gradient is Lipschitz, we have
		\begin{align*}
			f(\bm{w}^{\ddag})-f(\bm{w}^{\dag})\leq -\gamma\|\nabla f(\bm{w}^{\dag})\|+2\gamma\|\bm{m}-\nabla f(\bm{w}^{\dag})\|+\frac{L\gamma^2}{2}.
		\end{align*}
	\end{lemma}
	
	\section{Proof of Theorem \ref{th1}}
	We designate the following variables as
	\begin{equation}\label{th1-temp1}
		\bm{\epsilon}^t:=\bm{m}^t-\nabla f(\bm{w}^t), \bm{\delta}^t:=\nabla f(\bm{w}^t;\xi^t)-\nabla f(\bm{w}^t),\bm{s}^t:=\nabla f(\bm{w}^{t-1})-\nabla f(\bm{w}^{t}).
	\end{equation}
	Based on the scheme of momentum updating in  the NSGD,
	\begin{align*}
		\bm{m}^{t}&=\theta\bm{m}^{t-1}+(1-\theta)\nabla f(\bm{w}^t;\xi^t)\\
		&=\theta\Big[\bm{\epsilon}^{t-1}+\nabla f(\bm{w}^{t-1})\Big]+(1-\theta)\Big[\bm{\delta}^t+\nabla f(\bm{w}^t)\Big].
	\end{align*}
	The relation above then gives us
	\begin{align*}
		\bm{\epsilon}^t=\bm{m}^{t}-\nabla f(\bm{w}^t)=\theta\bm{\epsilon}^{t-1}+\theta\bm{s}^t+(1-\theta)\bm{\delta}^t.
	\end{align*}
	By employing mathematical induction, we can proceed to derive
	\begin{align*}
		\bm{\epsilon}^t= \theta^t\bm{\epsilon}^{0}+\sum_{k=1}^{t}\theta^{t-k}\bm{s}^k+(1-\theta)\sum_{k=1}^{t}\theta^{t-k}\bm{\delta}^k.
	\end{align*}
	By taking the norms of both sides of the inequality and considering the expectation,
	\begin{align}\label{th1-temp2}
		\begin{aligned}
			\EE\|\bm{\epsilon}^t\|&\leq \sum_{k=1}^{t}\theta^{t-k}\EE\|\bm{s}^k\|+(1-\theta)\EE\Big\|\sum_{k=1}^{t}\theta^{t-k}\bm{\delta}^k\Big\|+\theta^t\|\bm{\epsilon}^{0}\|\\
			&\leq \frac{L\gamma}{1-\theta} +(1-\theta)\EE\Big\|\sum_{k=1}^{t}\theta^{t-k}\bm{\delta}^k\Big\|+\theta^t\|\bm{\epsilon}^{0}\|,
		\end{aligned}
	\end{align}
	where we used the
	gradient Lipschitz property  of function $f$ as
	\begin{align*}
		\EE\|\bm{s}^k\|=\EE\|\nabla f(\bm{w}^{k})-\nabla f(\bm{w}^{k-1})\|\leq \EE\Big[L\|\bm{w}^{k}-\bm{w}^{k-1}\|\Big]\leq L\gamma.
	\end{align*}
	With the definition and Lemma \ref{lemma0}, we can get
	\begin{align*}
		\EE\Big\|\sum_{k=1}^{t}\theta^{t-k}\bm{\delta}^k\Big\|&\leq \sqrt{\EE\Big\|\sum_{k=1}^{t}\theta^{t-k}\bm{\delta}^k\Big\|^2}= \sqrt{\sum_{k=1}^{t}\theta^{2t-2k}\EE\|\bm{\delta}^k\|^2}\\
		&\leq \frac{\sqrt{4(B+L\gamma t)^{2-p}\sigma^p}}{\sqrt{1-\theta^2}}\leq\frac{2(B+L\gamma t)^{\frac{2 -p}{2}}\sigma^{p/2}}{\sqrt{1-\theta}}.
	\end{align*}
	The bound of the right side of  \eqref{th1-temp2} then reduces to
	\begin{align}\label{th1-temp3}
		\begin{aligned}
			\EE\|\bm{\epsilon}^t\|&\leq \frac{L\gamma}{1-\theta} +2\sqrt{1-\theta}(B+L\gamma t)^{\frac{2 -p}{2}}\sigma^{p/2}+\theta^t\|\bm{\epsilon}^{0}\|\\
			&\leq \frac{L\gamma}{1-\theta} +2\sqrt{1-\theta}(B+L\gamma T)^{\frac{2 -p}{2}}\sigma^{p/2}+\theta^t\|\bm{\epsilon}^{0}\|.
		\end{aligned}
	\end{align}
	Summing the inequality from $t=1$ to $T$, we then get
	\begin{align}\label{th1-temp4}
		\begin{aligned}
			\sum_{t=1}^{T}\EE\|\bm{\epsilon}^t\|/T
			&\leq\frac{ L\gamma}{1-\theta} +\frac{\|\nabla f(\bm{w}^0)\|}{(1-\theta)T}+2\sqrt{1-\theta}(B+L\gamma T)^{\frac{2 -p}{2}}\sigma^{p/2}.
		\end{aligned}
	\end{align}
		Applying Lemma \ref{lemma1} with $\bm{w}^{\dag}$ replaced by $\bm{w}^{t}$ and $\bm{m}$ replaced by $\bm{m}^{t}$, and  considering expectations,
	\begin{align*}
		\EE f(\bm{w}^{t+1})-\EE f(\bm{w}^{t})\leq -\gamma \EE\|\nabla f(\bm{w}^{t})\|+2\gamma\EE\|\bm{\epsilon}^t\|+\frac{L\gamma^2}{2}.
	\end{align*}
	Adding up the recursion from $t=1$ to $T$, we obtain
	\begin{align}\label{conver-core}
		\frac{1}{T}\sum_{t=1}^T\EE\|\nabla f(\bm{w}^{t})\|&\leq \frac{f(\bm{w}^{1})-\min f}{\gamma T}+  2\sum_{t=1}^T\EE\|\bm{\epsilon}^t\|/T
		+L\gamma/2.
	\end{align}
Employing the bound \eqref{th1-temp4} and \eqref{conver-core}, we further get
	\begin{align}\label{th1-bound}
		\begin{aligned}
			&\frac{1}{T}\sum_{t=1}^T\EE\|\nabla f(\bm{w}^{t})\|\\
			&\leq \frac{f(\bm{w}^{1})-\min f}{\gamma T}+
			L\gamma/2+\frac{2L\gamma}{1-\theta} +\frac{2\|\nabla f(\bm{w}^0)\|}{(1-\theta)T}+4\sqrt{1-\theta}(B+L\gamma T)^{\frac{2 -p}{2}}\sigma^{p/2}\\
			&\leq   \frac{f(\bm{w}^{1})-\min f}{\gamma T}+\frac{3L\gamma}{1-\theta} +\frac{2\|\nabla f(\bm{w}^0)\|}{(1-\theta)T}+4\sqrt{1-\theta}(B+L\gamma T)^{\frac{2 -p}{2}}\sigma^{p/2}\\
			&\leq \frac{f(\bm{w}^{1})-\min f}{\gamma T}+\frac{3L\gamma}{1-\theta} +\frac{2\|\nabla f(\bm{w}^0)\|}{(1-\theta)T}+4(B+L)^{\frac{2 -p}{2}}\sqrt{1-\theta}(\gamma T)^{\frac{2 -p}{2}}\sigma^{p/2},
		\end{aligned}
	\end{align}
	where we used $	L\gamma/2\leq \frac{L\gamma}{1-\theta}$ and $(B+L\gamma T)\leq (B+L)\gamma T$ because $\gamma T\geq 1$ (Note that we have $\frac{f(\bm{w}^{1})-\min f}{\gamma T}$ in the bound, $\gamma T\rightarrow +\infty$ to ensures the convergence).

\medskip
	 {\color{black}\textbf{Case I: $L\Delta\geq 1$.}}
	\noindent 1) If $\sigma^{\frac{4p-4}{3p-2}}T^{\frac{p}{3p-2}}\geq {\color{black}\sqrt{L\Delta}}$, we have
	$$1-\theta=\frac{{\color{black}\sqrt{L\Delta}}}{\sigma^{\frac{4p-4}{3p-2}}T^{\frac{p}{3p-2}}},\gamma=\frac{{\color{black}\sqrt{\Delta}^{\frac{3}{4}}}}{{\color{black}\sqrt[4]{L}}\sigma^{\frac{2p-2}{3p-2}}T^{\frac{2p-1}{3p-2}}}.$$
	Based on \eqref{th1-bound} and the fact $\bm{w}^0=\bm{w}^1$, we have
	\begin{align}\label{th1-b1}
		\begin{aligned}
			&\frac{1}{T}\sum_{t=1}^T\EE\|\nabla f(\bm{w}^{t})\|\leq \frac{{\color{black}4(L\Delta)^{\frac{1}{4}}}\sigma^{\frac{2p-2}{3p-2}}}{T^{\frac{p-1}{3p-2}}}+\frac{2\|\nabla f(\bm{w}^0)\|\sigma^{\frac{4p-4}{3p-2}}}{T^{\frac{2p-2}{3p-2}}}\\
			&\qquad+4{\color{black}(L\Delta)^{\frac{6-3p}{8}}}\Big(\frac{B}{L}+1\Big)^{\frac{2-p}{2}}\sigma^{\frac{5p^2-12p+8}{6p-4}}\frac{1}{T^{\frac{p}{2(3p-2)}-\frac{(p-1)(2-p)}{2(3p-2)}}}\\
			&\quad\leq \frac{{\color{black}4(L\Delta)^{\frac{1}{4}}}\sigma^{\frac{2p-2}{3p-2}}}{T^{\frac{p-1}{3p-2}}}+\frac{2\|\nabla f(\bm{w}^0)\|\sigma^{\frac{4p-4}{3p-2}}}{T^{\frac{2p-2}{3p-2}}}+\frac{4{\color{black}(L\Delta)^{\frac{6-3p}{8}}}\Big(\frac{B}{L}+1\Big)^{\frac{2-p}{2}}\sigma^{\frac{5p^2-12p+8}{6p-4}}}{T^{\frac{(p-2)^2}{2(3p-2)}+\frac{p-1}{3p-2}}},
		\end{aligned}
	\end{align}
	where we used $1<p\leq 2$.
	
	\noindent 2) If $\sigma^{\frac{4p-4}{3p-2}}T^{\frac{p}{3p-2}}\leq 1$, we have
	$$1-\theta=1,\gamma=\frac{{\color{black}\sqrt{\Delta}}}{{\color{black}\sqrt{L}}T^{\frac{1}{2}}}.$$
	The upper bound \eqref{th1-bound} can be rebounded as
	\begin{align*}
		&\frac{1}{T}\sum_{t=1}^T\EE\|\nabla f(\bm{w}^{t})\|\\
        &\leq \frac{{\color{black}4\sqrt{L\Delta}}}{T^{\frac{1}{2}}}+\frac{2\|\nabla f(\bm{w}^0)\|}{T}+4{\color{black}(L\Delta)^{\frac{2-p}{4}}}\Big(\frac{B}{L}+1\Big)^{\frac{2-p}{2}}T^{\frac{2 -p}{4}}\sigma^{p/2}\\
		&\leq \frac{{\color{black}4\sqrt{L\Delta}}}{T^{\frac{1}{2}}}+\frac{2\|\nabla f(\bm{w}^0)\|}{T}+4{\color{black}(L\Delta)^{\frac{2-p}{4}}}\Big(\frac{B}{L}+1\Big)^{\frac{2-p}{2}}\frac{\sigma^{p/8}}{T^{\frac{3p^2}{32p-32}-\frac{2-p}{4}}}.
	\end{align*}
	Notice that $\frac{3p^2}{32p-32}-\frac{2-p}{4}\geq\frac{1}{4}$,
	when $1<p\leq 2$. Thus, we are led to
	\begin{align}\label{th1-b2}
		\begin{aligned}
			\frac{1}{T}\sum_{t=1}^T\EE\|\nabla f(\bm{w}^{t})\|\leq \frac{4{\color{black}(L\Delta)^{\frac{2-p}{4}}}\Big(\frac{B}{L}+1\Big)^{\frac{2-p}{2}}\sigma^{p/8}}{T^{\frac{1}{4}}}+\frac{{\color{black}4\sqrt{L\Delta}}}{T^{\frac{1}{2}}}+\frac{2\|\nabla f(\bm{w}^0)\|}{T}.
		\end{aligned}
	\end{align}
	Taking maximum of the both bounds in \eqref{th1-b1} and \eqref{th1-b2}, we then get the convergence as
	\begin{align*}
		\begin{aligned}
			\frac{1}{T}\sum_{t=1}^T\EE\|\nabla f(\bm{w}^{t})\|&\leq \frac{{\color{black}4(L\Delta)^{\frac{1}{4}}}\sigma^{\frac{2p-2}{3p-2}}}{T^{\frac{p-1}{3p-2}}}+\frac{2\|\nabla f(\bm{w}^0)\|\sigma^{\frac{4p-4}{3p-2}}}{T^{\frac{2p-2}{3p-2}}}\\
&+\frac{4{\color{black}(L\Delta)^{\frac{6-3p}{8}}}\Big(\frac{B}{L}+1\Big)^{\frac{2-p}{2}}\sigma^{\frac{5p^2-12p+8}{6p-4}}}{T^{\frac{(p-2)^2}{2(3p-2)}+\frac{p-1}{3p-2}}}\\
			&+\frac{4{\color{black}(L\Delta)^{\frac{2-p}{4}}}\Big(\frac{B}{L}+1\Big)^{\frac{2-p}{2}}\sigma^{p/8}}{T^{\frac{1}{4}}}+\frac{{\color{black}4\sqrt{L\Delta}}}{T^{\frac{1}{2}}}+\frac{2\|\nabla f(\bm{w}^0)\|}{T}.
		\end{aligned}
	\end{align*}

	 {\color{black}\textbf{Case II: $L\Delta<1$.} Similar to Case I, we can get
     	\begin{align*}
		\begin{aligned}
			\frac{1}{T}\sum_{t=1}^T\EE\|\nabla f(\bm{w}^{t})\|&\leq \frac{{\color{black}4\sqrt{L\Delta}}\sigma^{\frac{2p-2}{3p-2}}}{T^{\frac{p-1}{3p-2}}}+\frac{2\|\nabla f(\bm{w}^0)\|\sigma^{\frac{4p-4}{3p-2}}}{T^{\frac{2p-2}{3p-2}}}\\
&+\frac{4{\color{black}(L\Delta)^{\frac{2-p}{4}}}\Big(\frac{B}{L}+1\Big)^{\frac{2-p}{2}}\sigma^{\frac{5p^2-12p+8}{6p-4}}}{T^{\frac{(p-2)^2}{2(3p-2)}+\frac{p-1}{3p-2}}}\\
			&+\frac{4{\color{black}(L\Delta)^{\frac{2-p}{4}}}\Big(\frac{B}{L}+1\Big)^{\frac{2-p}{2}}\sigma^{p/8}}{T^{\frac{1}{4}}}+\frac{{\color{black}4\sqrt{L\Delta}}}{T^{\frac{1}{2}}}+\frac{2\|\nabla f(\bm{w}^0)\|}{T}\\
            &\leq \frac{{\color{black}4(L\Delta)^{\frac{1}{4}}}\sigma^{\frac{2p-2}{3p-2}}}{T^{\frac{p-1}{3p-2}}}+\frac{2\|\nabla f(\bm{w}^0)\|\sigma^{\frac{4p-4}{3p-2}}}{T^{\frac{2p-2}{3p-2}}}\\
&+\frac{4{\color{black}(L\Delta)^{\frac{2-p}{4}}}\Big(\frac{B}{L}+1\Big)^{\frac{2-p}{2}}\sigma^{\frac{5p^2-12p+8}{6p-4}}}{T^{\frac{(p-2)^2}{2(3p-2)}+\frac{p-1}{3p-2}}}\\
			&+\frac{4{\color{black}(L\Delta)^{\frac{2-p}{4}}}\Big(\frac{B}{L}+1\Big)^{\frac{2-p}{2}}\sigma^{p/8}}{T^{\frac{1}{4}}}+\frac{{\color{black}4\sqrt{L\Delta}}}{T^{\frac{1}{2}}}+\frac{2\|\nabla f(\bm{w}^0)\|}{T}
		\end{aligned}
	\end{align*}
     as $L\Delta<1$.

Combining Case I and Case II, we then proved the final result.

}


	\section{Proof of Theorem \ref{th2}}
	For Algorithm \ref{alg2}, we use different shorthand notation
	\begin{equation}\label{th2-temp1}
		\begin{aligned}
			\bm{\epsilon}^t:=\bm{m}^t-\nabla f(\bm{w}^t), \bm{\delta}^t:=\nabla f(\bm{w}^t;\xi^t)-\nabla f(\bm{w}^t),\\
			\hat{\bm{s}}^t:=\nabla f(\bm{w}^{t-1})-\nabla f(\bm{w}^{t})+\nabla f(\bm{w}^{t};\xi^t)-\nabla f(\bm{w}^{t-1};\xi^t).
		\end{aligned}
	\end{equation}
	The scheme of momentum updating in  the NSGD-VR indicates
	\begin{align*}
		\bm{m}^{t}&=\theta\bm{m}^{t-1}+\nabla f(\bm{w}^t;\xi^t)-\theta\nabla f(\bm{w}^{t-1};\xi^t)\\
		&=\theta\Big[\bm{\epsilon}^{t-1}+\nabla f(\bm{w}^{t-1})\Big]+(1-\theta)\Big[\bm{\delta}^t+\nabla f(\bm{w}^t)\Big]+\theta\Big[\nabla f(\bm{w}^{t};\xi^t)-\nabla f(\bm{w}^{t-1};\xi^t)\Big].
	\end{align*}
	Thus, we can derive
	\begin{align*}
		\bm{\epsilon}^t=\bm{m}^t-\nabla f(\bm{w}^t)=\theta\bm{\epsilon}^{t-1}+\theta\hat{\bm{s}}^t+(1-\theta)\bm{\delta}^t.
	\end{align*}
	The relation above then gives us
	\begin{align*}
		\bm{\epsilon}^t= \theta^t\bm{\epsilon}^{0}+\sum_{k=1}^{t}\theta^{t-k}\hat{\bm{s}}^k+(1-\theta)\sum_{k=1}^{t}\theta^{t-k}\bm{\delta}^k.
	\end{align*}
	Taking norms of both sides and using the triangle inequality, it follows
	\begin{align*}
		\|\bm{\epsilon}^t\|\leq  \theta^t\|\bm{\epsilon}^{0}\|+\|\sum_{k=1}^{t}\theta^{t-k}\hat{\bm{s}}^k\|+(1-\theta)\|\sum_{k=1}^{t}\theta^{t-k}\bm{\delta}^k\|.
	\end{align*}
	Taking the expectations, we can further have
	\begin{align*}
		\EE\|\bm{\epsilon}^t\|\leq  \theta^t\EE\|\bm{\epsilon}^{0}\|+\EE\|\sum_{k=1}^{t}\theta^{t-k}\hat{\bm{s}}^k\|+(1-\theta)\EE\|\sum_{k=1}^{t}\theta^{t-k}\bm{\delta}^k\|.
	\end{align*}
	The upper bound of $\|\sum_{k=1}^{t}\theta^{t-k}\hat{\bm{s}}^k\|$ can be derived as follows
	\begin{align*}
		\EE\|\sum_{k=1}^{t}\theta^{t-k}\hat{\bm{s}}^k\|&\leq\sqrt{\EE\|\sum_{k=1}^{t}\theta^{t-k}\hat{\bm{s}}^k\|^2}=\sqrt{\sum_{k=1}^{t}\theta^{2t-2k}\EE\|\hat{\bm{s}}^k\|^2}\\
		&\leq\sqrt{\frac{4L^2\gamma^2}{1-\theta^2}}\leq\frac{2L\gamma}{\sqrt{1-\theta}},
	\end{align*}
	where we used $\EE\hat{\bm{s}}^k=\bm{0}$ and
	\begin{align*}
		\EE\|\hat{\bm{s}}^k\|^2&\leq 2\EE\|\nabla f(\bm{w}^{t-1})-\nabla f(\bm{w}^{t})\|^2+2\|\nabla f(\bm{w}^{t};\xi^t)-\nabla f(\bm{w}^{t-1};\xi^t)\|^2\\
		&\leq 4L^2\EE\|\bm{w}^{t}-\bm{w}^{t-1}\|^2\leq 4L^2\gamma^2.
	\end{align*}
	The bound of $\EE\Big\|\sum_{k=1}^{t}\theta^{t-k}\bm{\delta}^k\Big\|$ follows the same of that in Theorem \ref{th1}. Therefore, we are then led to the following result
	\begin{align}\label{th2-temp2}
		\begin{aligned}
			\EE\|\bm{\epsilon}^t\|&\leq \frac{2L\gamma}{\sqrt{1-\theta}} +2\sqrt{1-\theta}(B+L\gamma T)^{\frac{2 -p}{2}}\sigma^{p/2}+\theta^t\|\bm{\epsilon}^{0}\|
		\end{aligned}
	\end{align}
	as $1\leq t\leq T$. Adding up the inequality from $t=1$ to $T$, we obtain
	\begin{align}\label{th2-temp3}
		\begin{aligned}
			\sum_{t=1}^{T}\EE\|\bm{\epsilon}^t\|/T
			&\leq\frac{2L\gamma}{\sqrt{1-\theta}} +\frac{\|\nabla f(\bm{w}^0)\|}{(1-\theta)T}+2\sqrt{1-\theta}(B+L\gamma T)^{\frac{2 -p}{2}}\sigma^{p/2}.
		\end{aligned}
	\end{align}
	Noticing that \eqref{conver-core} still holds in NSGD-VR, we then have
	\begin{align}\label{th2-bound}
		\begin{aligned}
			&\frac{1}{T}\sum_{t=1}^T\EE\|\nabla f(\bm{w}^{t})\|\\
			&\leq \frac{f(\bm{w}^{1})-\min f}{\gamma T}+
			L\gamma/2+\frac{4L\gamma}{\sqrt{1-\theta}} +\frac{2\|\nabla f(\bm{w}^0)\|}{(1-\theta)T}+4\sqrt{1-\theta}(B+L\gamma T)^{\frac{2 -p}{2}}\sigma^{p/2}\\
			&\leq   \frac{f(\bm{w}^{1})-\min f}{\gamma T}+\frac{5L\gamma}{\sqrt{1-\theta}} +\frac{2\|\nabla f(\bm{w}^0)\|}{(1-\theta)T}+4\sqrt{1-\theta}(B+L\gamma T)^{\frac{2 -p}{2}}\sigma^{p/2}\\
			&\leq \frac{f(\bm{w}^{1})-\min f}{\gamma T}+\frac{5L\gamma}{\sqrt{1-\theta}} +\frac{2\|\nabla f(\bm{w}^0)\|}{(1-\theta)T}+4(B+L)^{\frac{2 -p}{2}}\sqrt{1-\theta}(\gamma T)^{\frac{2 -p}{2}}\sigma^{p/2},
		\end{aligned}
	\end{align}
	where we used $	L\gamma/2\leq \frac{L\gamma}{\sqrt{1-\theta}}$ and $(B+L\gamma T)\leq (B+L)\gamma T$.
	
	\medskip
	
	\noindent	1). If $\sigma^{\frac{p}{2p-1}}T^{\frac{p}{2p-1}}\geq 1$, it follows
	$$1-\theta=\frac{1}{\sigma^{\frac{p}{2p-1}}T^{\frac{p}{2p-1}}},\gamma=\frac{1}{L\sigma^{\frac{p}{4(2p-1)}}T^{\frac{5p-2}{4(2p-1)}}}.$$
	We then simplify the inequality \eqref{th2-bound} as
	\begin{align}\label{th2-b1-pre}
		\begin{aligned}
			&\frac{1}{T}\sum_{t=1}^T\EE\|\nabla f(\bm{w}^{t})\|\\
			&\leq \frac{f(\bm{w}^{1})-\min f}{\gamma T}+\frac{5L\gamma}{\sqrt{1-\theta}} +\frac{2\|\nabla f(\bm{w}^0)\|}{(1-\theta)T}+4(B+L)^{\frac{2 -p}{2}}\sqrt{1-\theta}(\gamma T)^{\frac{2 -p}{2}}\sigma^{p/2}\\
			&\leq\frac{L(f(\bm{w}^{1})-\min f){\color{black}\sigma^{\frac{p}{4(2p-1)}}}+5{\color{black}\sigma^{\frac{p}{4(2p-1)}}}}{T^{\frac{3p-2}{4(2p-1)}}} +\frac{2\|\nabla f(\bm{w}^0)\|{\color{black}\sigma^{\frac{p}{2p-1}}}}{T^{\frac{p-1}{2p-1}}}+\frac{4\Big(\frac{B}{L}+1\Big)^{\frac{2 -p}{2}}{\color{black}\sigma^{\frac{9p^2-10p}{8(2p-1)}}} }{{\color{black}T^{\frac{3p^2-4p+4}{8(2p-1)}}}}\\
			&=\frac{L(f(\bm{w}^{1})-\min f){\color{black}\sigma^{\frac{p}{4(2p-1)}}}+5{\color{black}\sigma^{\frac{p}{4(2p-1)}}}}{T^{\frac{3p-2}{4(2p-1)}}} +\frac{2\|\nabla f(\bm{w}^0)\|{\color{black}\sigma^{\frac{p}{2p-1}}}}{T^{\frac{p-1}{2p-1}}}+\frac{4\Big(\frac{B}{L}+1\Big)^{\frac{2 -p}{2}}{\color{black}\sigma^{\frac{9p^2-10p}{8(2p-1)}}} }{{\color{black}T^{\frac{p-1}{2p-1}+\frac{3(p-2)^2}{8(2p-1)}}}}\\
&\leq\frac{L(f(\bm{w}^{1})-\min f){\color{black}\sigma^{\frac{p}{4(2p-1)}}}+5{\color{black}\sigma^{\frac{p}{4(2p-1)}}}}{T^{\frac{3p-2}{4(2p-1)}}} +\frac{2\|\nabla f(\bm{w}^0)\|{\color{black}\sigma^{\frac{p}{2p-1}}}}{T^{\frac{p-1}{2p-1}}}\\
&\qquad+\frac{4\Big(\frac{B}{L}+1\Big)^{\frac{2 -p}{2}}{\color{black}\sigma^{\frac{12p^2-22p+12}{8(2p-1)}}} }{{\color{black}T^{\frac{p-1}{2p-1}}}},
		\end{aligned}
	\end{align}
{\color{black}due to   $\frac{1}{T}\leq \sigma$.}
	Based on the facts above and $\bm{w}^0=\bm{w}^1$,
	\begin{align}\label{th2-b1}
		\begin{aligned}
			&\frac{1}{T}\sum_{t=1}^T\EE\|\nabla f(\bm{w}^{t})\|\\
			&\leq\frac{2\|\nabla f(\bm{w}^0)\|{\color{black}\sigma^{\frac{p}{2p-1}}}+4\Big(\frac{B}{L}+1\Big)^{\frac{2 -p}{2}}{\color{black}\sigma^{\frac{12p^2-22p+12}{8(2p-1)}}}}{T^{\frac{p-1}{2p-1}}}
			+\frac{[L(f(\bm{w}^{1})-\min f)+5]{\color{black}\sigma^{\frac{p}{4(2p-1)}}}}{T^{\frac{3p-2}{4(2p-1)}}}\\
			&\leq\frac{2\|\nabla f(\bm{w}^0)\|{\color{black}\sigma^{\frac{p}{2p-1}}}+4\Big(\frac{B}{L}+1\Big)^{\frac{2 -p}{2}}{\color{black}\sigma^{\frac{12p^2-22p+12}{8(2p-1)}}}+[L(f(\bm{w}^{1})-\min f)+5]{\color{black}\sigma^{\frac{p}{4(2p-1)}}}}{T^{\frac{p-1}{2p-1}}}
		\end{aligned}
	\end{align}
	because $\frac{3p-2}{4(2p-1)}\geq \frac{p-1}{2p-1}$.
	
	\noindent	2). If ${\color{black}\sigma^{\frac{p}{2p-1}}}T^{\frac{p}{2p-1}}\leq 1$, it follows
	$$1-\theta=1,\gamma=\frac{1}{LT^{1/2}}.$$
	We can rebound   \eqref{th2-bound} as
	\begin{align}\label{th2-b2-pre}
		\begin{aligned}
			\frac{1}{T}\sum_{t=1}^T\EE\|\nabla f(\bm{w}^{t})\|
			&\leq \frac{L(f(\bm{w}^{1})-\min f)+5}{ \sqrt{T}}+\frac{2\|\nabla f(\bm{w}^0)\|}{T}+\frac{4\Big(\frac{B}{L}+1\Big)^{\frac{2 -p}{2}}}{T^{\frac{p^2}{2p-1}-\frac{2-p}{4}}}.
		\end{aligned}
	\end{align}
	It is easy to check that
	$$\frac{p^2}{2p-1}-\frac{2-p}{4}\geq\frac{1}{2}$$
	as $p\in(1,2]$.
	The bound in \eqref{th2-b2-pre} can be rebounded as
	\begin{align}\label{th2-b2}
		\begin{aligned}
			\frac{1}{T}\sum_{t=1}^T\EE\|\nabla f(\bm{w}^{t})\|
			&\leq \frac{L(f(\bm{w}^{1})-\min f)+4\Big(\frac{B}{L}+1\Big)^{\frac{2 -p}{2}}+5}{ \sqrt{T}}+\frac{2\|\nabla f(\bm{w}^0)\|}{T}\\
			&=\frac{L(f(\bm{w}^{0})-\min f)+4\Big(\frac{B}{L}+1\Big)^{\frac{2 -p}{2}}+5}{ \sqrt{T}}+\frac{2\|\nabla f(\bm{w}^0)\|}{T}.
		\end{aligned}
	\end{align}
	Summing both bounds in \eqref{th2-b1} and \eqref{th2-b2}, the convergence rate is proved as
	\begin{align*}
		\begin{aligned}
			&\frac{1}{T}\sum_{t=1}^T\EE\|\nabla f(\bm{w}^{t})\|\\
			&\leq \frac{2\|\nabla f(\bm{w}^0)\|{\color{black}\sigma^{\frac{p}{2p-1}}}+4\Big(\frac{B}{L}+1\Big)^{\frac{2 -p}{2}}{\color{black}\sigma^{\frac{12p^2-22p+12}{8(2p-1)}}}+[L(f(\bm{w}^{0})-\min f)+5]{\color{black}\sigma^{\frac{p}{4(2p-1)}}}}{T^{\frac{p-1}{2p-1}}}\\
			&\qquad+\frac{L(f(\bm{w}^{0})-\min f)+4\Big(\frac{B}{L}+1\Big)^{\frac{2 -p}{2}}+5}{ \sqrt{T}}+\frac{2\|\nabla f(\bm{w}^0)\|}{T}.
		\end{aligned}
	\end{align*}
    Note the following inequality always holds
\begin{align}\label{sigin2}
    \sigma^{y} \leq \sigma^a + \sigma^{x},~\textrm{as}~0 \leq x \leq y\leq a.
\end{align}
This holds because if \(\sigma \geq 1\), then \(\sigma^{y} \leq \sigma^a\); otherwise, \(\sigma^{y} \leq \sigma^{x}\). By combining these cases, the inequality is established. 
Due to that $\frac{p}{4(2p-1)}\leq \frac{12p^2-22p+12}{8(2p-1)}\leq\frac{p}{2p-1}$ when $1\leq p\leq 2$, using \eqref{sigin2}, we then proved the result.
	
	\section{Proof of Theorem \ref{th1-c}}
	This proof denotes the following shorthand notation
	\begin{equation}\label{th1-c-temp1}
		\begin{aligned}
			&		\bm{\epsilon}^t:=\bm{m}^t-\nabla f(\bm{w}^t), \bm{\delta}^t:=\Ch_{h}(\bm{g}^t)-\EE\Ch_{h}(\bm{g}^t),\\
			&\bm{c}^t:=\EE\Ch_{h}(\bm{g}^t)-\nabla f(\bm{w}^t),\bm{s}^t:=\nabla f(\bm{w}^{t-1})-\nabla f(\bm{w}^{t}).
		\end{aligned}
	\end{equation}
	In NSGDC, we have
	\begin{align*}
		\bm{m}^{t}&=\theta\bm{m}^{t-1}+(1-\theta)\Ch_{h}(\bm{g}^t)\\
		&=\theta\Big[\bm{\epsilon}^{t-1}+\nabla f(\bm{w}^{t-1})\Big]+(1-\theta)\Big[\bm{\delta}^t+\bm{c}^t+\nabla f(\bm{w}^t)\Big].
	\end{align*}
	The inequality then yields
	\begin{align*}
		\bm{\epsilon}^t=\bm{m}^{t}-\nabla f(\bm{w}^t)=\theta\bm{\epsilon}^{t-1}+\theta\bm{s}^t+(1-\theta)\bm{\delta}^t+(1-\theta)\bm{c}^t.
	\end{align*}
	Using mathematical induction,  we can derive
	\begin{align*}
		\bm{\epsilon}^t= \theta^t\bm{\epsilon}^{0}+\sum_{k=1}^{t}\theta^{t-k}\bm{s}^k+(1-\theta)\sum_{k=1}^{t}\theta^{t-k}\bm{\delta}^k+(1-\theta)\sum_{k=1}^{t}\theta^{t-k}\bm{c}^k.
	\end{align*}
	By taking the norms  and expectations  of both sides, we have
	\begin{align}\label{th1-c-temp2}
		\begin{aligned}
			\EE\|\bm{\epsilon}^t\|&\leq \sum_{k=1}^{t}\theta^{t-k}\EE\|\bm{s}^k\|+(1-\theta)\sum_{k=1}^{t}\theta^{t-k}\EE\|\bm{c}^k\|+(1-\theta)\EE\Big\|\sum_{k=1}^{t}\theta^{t-k}\bm{\delta}^k\Big\|+\theta^t\|\bm{\epsilon}^{0}\|\\
			&\leq \frac{L\gamma}{1-\theta} +2\sigma^{p}h^{-(p-1)}+(1-\theta)\EE\Big\|\sum_{k=1}^{t}\theta^{t-k}\bm{\delta}^k\Big\|+\theta^t\|\bm{\epsilon}^{0}\|,
		\end{aligned}
	\end{align}
	where we used the
	gradient Lipschitz property  of function $f$ as
	\begin{align*}
		\EE\|\bm{s}^k\|=\EE\|\nabla f(\bm{w}^{k})-\nabla f(\bm{w}^{k-1})\|\leq \EE\Big[L\|\bm{w}^{k}-\bm{w}^{k-1}\|\Big]\leq L\gamma,
	\end{align*}
	and Lemma \ref{lemma-c} (note that $h\geq 2(\gamma LT+\|\nabla f(\bm{w}^0)\|)$). Now, we turn to the bound of $\EE\Big\|\sum_{k=1}^{t}\theta^{t-k}\bm{\delta}^k\Big\|$. With the Cauchy's inequality, we can get
	\begin{align*}
		\EE\Big\|\sum_{k=1}^{t}\theta^{t-k}\bm{\delta}^k\Big\|&\leq \sqrt{\EE\Big\|\sum_{k=1}^{t}\theta^{t-k}\bm{\delta}^k\Big\|^2}= \sqrt{\sum_{k=1}^{t}\theta^{2t-2k}\EE\|\bm{\delta}^k\|^2}\\
		&\leq \frac{\sqrt{ 10h^{2-p}\sigma^{p}}}{\sqrt{1-\theta^2}}\leq\frac{4h^{\frac{2 -p}{2}}\sigma^{p/2}}{\sqrt{1-\theta}}.
	\end{align*}
	Recalling inequality \eqref{th1-c-temp2}, we are led to
	\begin{align*} 
		\begin{aligned}
			\EE\|\bm{\epsilon}^t\|&\leq \frac{L\gamma}{1-\theta}  +2\sigma^{p}h^{-(p-1)}+4\sqrt{1-\theta}h^{\frac{2 -p}{2}}\sigma^{p/2}+\theta^t\|\bm{\epsilon}^{0}\|,
		\end{aligned}
	\end{align*}
	which further indicates
	\begin{align}\label{th1-c-temp3}
		\begin{aligned}
			\sum_{t=1}^T\EE\|\bm{\epsilon}^t\|/T&\leq \frac{L\gamma}{1-\theta}  +2\sigma^{p}h^{-(p-1)}+4\sqrt{1-\theta}h^{\frac{2 -p}{2}}\sigma^{p/2} +\frac{\|\nabla f(\bm{w}^0)\|}{(1-\theta)T}.
		\end{aligned}
	\end{align}
 Noticing that  \eqref{conver-core} still holds,
further with \eqref{th1-c-temp3},  we can get the following bound
	\begin{align}\label{th1-c-bound}
		\begin{aligned}
			&\frac{1}{T}\sum_{t=1}^T\EE\|\nabla f(\bm{w}^{t})\|\\
			&\leq \frac{f(\bm{w}^{1})-\min f}{\gamma T}+
			L\gamma/2+\frac{2L\gamma}{1-\theta} +\frac{2\|\nabla f(\bm{w}^0)\|}{(1-\theta)T}+4\sigma^{p}h^{-(p-1)}+8\sqrt{1-\theta}h^{\frac{2 -p}{2}}\sigma^{p/2}\\
			&\leq   \frac{f(\bm{w}^{1})-\min f}{\gamma T}+\frac{3L\gamma}{1-\theta} +\frac{2\|\nabla f(\bm{w}^0)\|}{(1-\theta)T}+4\sigma^{p}h^{-(p-1)}+8\sqrt{1-\theta}h^{\frac{2 -p}{2}}\sigma^{p/2},
		\end{aligned}
	\end{align}
	where we used that $\frac{1}{2}\leq\frac{1}{1-\theta}$.
	\medskip
    
{\color{black}
\textbf{Case I}: $L\Delta \geq 1$. 
We used a shorthand notation  as
$\zeta:=(L\Delta)^{\frac{p}{3p-2}}\geq 1.$}

	\noindent 1)  If $\sigma< 1$ and $\sigma^{\frac{4p-4}{3p-2}}T^{\frac{p}{3p-2}}\geq {\color{black}\zeta}$, we have
	$$1-\theta=\frac{{\color{black}\zeta}}{\sigma^{\frac{4p-4}{3p-2}}T^{\frac{p}{3p-2}}},\gamma=\frac{{\color{black} \sqrt{\zeta\Delta}}}{{\color{black}\sqrt{L}}\sigma^{\frac{2p-2}{3p-2}}T^{\frac{2p-1}{3p-2}}}, h=\frac{2{\color{black} \sqrt{\zeta}\sqrt{L\Delta}}T^{\frac{1}{3p-2}}}{{\color{black}\sigma^{\frac{2p-2}{3p-2}}}}+2\|\nabla f(\bm{w}^{0})\|.$$
{\color{black}Direct calculations indicate $\sqrt{\zeta}\sqrt{L\Delta}=\zeta^{\frac{2p-1}{p}}\geq 1$.}
	Based on \eqref{th1-c-bound},   with the fact $0\leq \sigma\leq 1$ and $\bm{w}^0=\bm{w}^1$,  we have
	\begin{align}\label{th1-c-b1}
		\begin{aligned}
			&\frac{1}{T}\sum_{t=1}^T\EE\|\nabla f(\bm{w}^{t})\|\leq \frac{{\color{black} (L\Delta)^{\frac{p-1}{3p-2}}}\sigma^{\frac{2p-2}{3p-2}}+3{\color{black} (L\Delta)^{\frac{p-1}{3p-2}}}\sigma^{\frac{2p-2}{3p-2}}+{\color{black}4\sigma^{\frac{p}{3p-2}}}}{T^{\frac{p-1}{3p-2}}}\\
			&\qquad+16\sqrt{1-\theta}\Big(\frac{{\color{black} \zeta^{\frac{2p-1}{p}}}T^{\frac{1}{3p-2}}}{{\color{black}\sigma^{\frac{2p-2}{3p-2}}}}+\|\nabla f(\bm{w}^{0})\|\Big)^{\frac{2 -p}{2}}\sigma^{\frac{p}{2}}+\frac{2\|\nabla f(\bm{w}^0)\|\sigma^{\frac{4p-4}{3p-2}}}{T^{\frac{2p-2}{3p-2}}}\\
			&\leq \frac{4{\color{black} (L\Delta)^{\frac{p-1}{3p-2}}}\sigma^{\frac{2p-2}{3p-2}}+{\color{black}4\sigma^{\frac{p}{3p-2}}}+16{\color{black} \zeta^{\frac{6p-2p^2-2}{2p}}}\sigma^{\frac{5p^2-12p+8}{2(3p-2)}}}{T^{\frac{p-1}{3p-2}}}\\
	&\qquad+\frac{16{\color{black}\sqrt{\zeta}}\|\nabla f(\bm{w}^{0})\|^{\frac{2 -p}{2}}{\color{black}\sigma^{\frac{3p^2-6p+4}{2(3p-2)}}}}{T^{\frac{p}{6p-4}}}+\frac{2\|\nabla f(\bm{w}^0)\|{\color{black}\sigma^{\frac{4p-4}{3p-2}}}}{T^{\frac{2p-2}{3p-2}}}\\
			&\leq \frac{ 4{\color{black} (L\Delta)^{\frac{p-1}{3p-2}}}\sigma^{\frac{2p-2}{3p-2}}+{\color{black}4\sigma^{\frac{p}{3p-2}}}+16{\color{black} (L\Delta)^{\frac{p}{3p-2}}}\sigma^{\frac{p-1}{3p-2}}}{T^{\frac{p-1}{3p-2}}}\\
			&\qquad+\frac{16{\color{black}(L\Delta)^{\frac{p}{6p-4}}}\|\nabla f(\bm{w}^{0})\|^{\frac{2 -p}{2}}\sigma^{\frac{3p^2-6p+4}{2(3p-2)}}}{T^{\frac{p}{6p-4}}}+\frac{2\|\nabla f(\bm{w}^0)\|{\color{black}\sigma^{\frac{4p-4}{3p-2}}}}{T^{\frac{2p-2}{3p-2}}},
		\end{aligned}
	\end{align}
	where we used that $(x+y)^{\alpha}\leq x^{\alpha}+y^{\alpha}$ as $0<\alpha\leq 1$ and $x,y>0$\footnote{We present a straightforward proof for this inequality. Without loss of generality, we can assume $x \geq y$. We can rewrite the inequality as
		$
		\left( \frac{x}{y} + 1 \right)^{\alpha} \leq \left( \frac{x}{y} \right)^{\alpha} + 1.
		$
		By letting $a = \frac{x}{y}$, it suffices to prove
		$(a + 1)^{\alpha}-a^{\alpha}  \leq  1$ for $a \geq 1.$
		It is easy to see that $(a+1)^{\alpha}- a^{\alpha}=\alpha(a+z)^{-\alpha}$ with some $0\leq z\leq 1$. We then proved the result by noting $a\geq 1$ and $0<\alpha\leq 1$. }, 
and 
$$\frac{p-1}{3p-2}\leq \frac{5p^2-12p+8}{6p-4}, {\color{black}\frac{6p-2p^2-2}{6p-4}}\leq\frac{p}{3p-2}$$
as $1\leq p\leq 2$.

	\noindent 2)  If $\sigma< 1$ and $\sigma^{\frac{4p-4}{3p-2}}T^{\frac{p}{3p-2}}\leq {\color{black}\zeta}$, we have
	$$1-\theta=1, \gamma={\color{black}\sqrt{\frac{\Delta}{L}}}\cdot\frac{1}{T^{\frac{1}{2}}}, h=2({\color{black}\sqrt{L\Delta}}T^{\frac{p+2}{6p-4}}+\|\nabla f(\bm{w}^{0})\|).$$
	The upper bound \eqref{th1-bound} can be rebounded as
	\begin{align*}
		&\frac{1}{T}\sum_{t=1}^T\EE\|\nabla f(\bm{w}^{t})\|\\
		&\leq \frac{4{\color{black}\sqrt{L\Delta}}}{T^{\frac{1}{2}}}+\frac{2\|\nabla f(\bm{w}^0)\|}{T}+4\sigma^{p}h^{-(p-1)}+8h^{\frac{2 -p}{2}}\sigma^{p/2}\\
		&\leq \frac{4{\color{black}\sqrt{L\Delta}}}{T^{\frac{1}{2}}}+\frac{2\|\nabla f(\bm{w}^0)\|}{T}+4T^{-{\color{black}\frac{p^2}{4p-4}}-\frac{(p-1)(p+2)}{6p-4}}\\
		&\quad+16{\color{black}(L\Delta)^{\frac{1}{4}}}T^{-{\color{black}\frac{3p^2}{32p-32}}+\frac{(2-p)(p+2)}{12p-8}} \sigma^{\frac{p}{8}}+16\|\nabla f(\bm{w}^{0})\|^{\frac{2 -p}{2}}T^{-{\color{black}\frac{p^2}{8p-8}}}.
	\end{align*}
	Noting that ${\color{black}\frac{p^2}{4p-4}}+\frac{(p-1)(p+2)}{6p-4}\geq \frac{1}{2}$,   ${\color{black}\frac{3p^2}{32p-32}}-\frac{(2-p)(p+2)}{12p-8}\geq\frac{p-1}{3p-2}$, and ${\color{black}\frac{p^2}{8p-8}}\geq\frac{1}{2}$  as $1<p\leq2$, we then get the following bound
	\begin{align}\label{th1-c-b2}
		\begin{aligned}
		&	\frac{1}{T}\sum_{t=1}^T\EE\|\nabla f(\bm{w}^{t})\|\leq \frac{{\color{black}16(L\Delta)^{\frac{1}{4}}}\sigma^{\frac{p}{8}}}{T^{\frac{p-1}{3p-2}}}+\frac{{\color{black}4\sqrt{L\Delta}}+16\|\nabla f(\bm{w}^{0})\|^{\frac{2 -p}{2}}+{\color{black}4}}{T^{\frac{1}{2}}}+\frac{2\|\nabla f(\bm{w}^0)\|}{T}.
		\end{aligned}
	\end{align}

	\noindent   3) {\color{black}Consider the subcase   $\sigma\geq 1$ and $\sigma^{\frac{2p}{3p-2}}T^{\frac{p}{3p-2}}\geq \zeta$.  }  Hence,
 we have
	$$1-\theta=\frac{{\color{black}\zeta}}{\sigma^{\frac{2p}{3p-2}}T^{\frac{p}{3p-2}}},\gamma={\color{black}\sqrt{\frac{\zeta\Delta}{L}}}\frac{1}{\sigma^{\frac{p}{3p-2}}T^{\frac{2p-1}{3p-2}}}, h=2{\color{black} \sqrt{\zeta}\sqrt{L\Delta}}T^{\frac{1}{3p-2}}{\color{black}\sigma^{\frac{2p}{3p-2}}}+2\|\nabla f(\bm{w}^{0})\|.$$
Using \eqref{th1-c-bound}, we then get
	\begin{align}\label{th1-c-b3}
		\begin{aligned}
		&\frac{1}{T}\sum_{t=1}^T\EE\|\nabla f(\bm{w}^{t})\|\leq \frac{{\color{black} 4(L\Delta)^{\frac{p-1}{3p-2}}}\sigma^{\frac{p}{3p-2}}+4\sigma^{\frac{p}{3p-2}}}{T^{\frac{p-1}{3p-2}}}\\
			&\qquad+16\sqrt{1-\theta}\Big({\color{black} \sqrt{\zeta}\sqrt{L\Delta} }T^{\frac{1}{3p-2}}\sigma^{\frac{{\color{black}2p}}{3p-2}}+\|\nabla f(\bm{w}^{0})\|\Big)^{\frac{2 -p}{2}}\sigma^{\frac{p}{2}}+\frac{2\|\nabla f(\bm{w}^0)\|\sigma^{\frac{2p}{3p-2}}}{T^{\frac{2p-2}{3p-2}}}\\
			&\leq \frac{{\color{black} 4(L\Delta)^{\frac{p-1}{3p-2}}\sigma^{\frac{p}{3p-2}}+4}\sigma^{\frac{p}{3p-2}}+16{\color{black} (L\Delta)^{\frac{6p-2p^2-2}{6p-4}}\sigma^{\frac{p^2}{6p-4}}}}{T^{\frac{p-1}{3p-2}}}\\
		&\qquad+\frac{16{\color{black}\sqrt{\zeta}}\|\nabla f(\bm{w}^{0})\|^{\frac{2 -p}{2}}\sigma^{\frac{3p^2-4p}{2(3p-2)}}}{T^{\frac{p}{6p-4}}}+\frac{2\|\nabla f(\bm{w}^0)\|\sigma^{\frac{2p}{3p-2}}}{T^{\frac{2p-2}{3p-2}}}\\
			&\leq \frac{{\color{black} 4(L\Delta)^{\frac{p-1}{3p-2}}\sigma^{\frac{p}{3p-2}}+4}\sigma^{\frac{p}{3p-2}}+16{\color{black} (L\Delta)^{\frac{6p-2p^2-2}{6p-4}}\sigma^{\frac{p^2}{6p-4}}}}{T^{\frac{p-1}{3p-2}}}+\frac{16{\color{black}\sqrt{\zeta}}\|\nabla f(\bm{w}^{0})\|^{\frac{2 -p}{2}}{\color{black}\sigma^{\frac{3p^2-4p}{2(3p-2)}}}}{T^{\frac{p}{6p-4}}}\\
			&\qquad+\frac{2\|\nabla f(\bm{w}^0)\|\sigma^{\frac{2p}{3p-2}}}{T^{\frac{2p-2}{3p-2}}}\\
			&\leq \frac{{\color{black} 4(L\Delta)^{\frac{p-1}{3p-2}}\sigma^{\frac{p}{3p-2}}+4}\sigma^{\frac{p}{3p-2}}+16{\color{black} (L\Delta)^{\frac{6p-2p^2-2}{6p-4}}\sigma^{\frac{p^2}{6p-4}}}}{T^{\frac{p-1}{3p-2}}}\\
			&\qquad+\frac{16{\color{black}(L\Delta)^{\frac{p}{6p-4}}}\|\nabla f(\bm{w}^{0})\|^{\frac{2 -p}{2}}\sigma^{\frac{3p^2-6p+4}{2(3p-2)}}}{T^{\frac{p}{6p-4}}}+\frac{2\|\nabla f(\bm{w}^0)\|\sigma^{\frac{2p}{3p-2}}}{T^{\frac{2p-2}{3p-2}}},
		\end{aligned}
	\end{align}
where we used $\frac{3p^2-4p}{2(3p-2)}\leq\frac{3p^2-6p+4}{2(3p-2)}$ as $p\in(1,2]$.

 {\color{black}4) If $\sigma\geq 1$ and $\sigma^{\frac{2p}{3p-2}}T^{\frac{p}{3p-2}}< \zeta$,
 we have
	$$1-\theta=1, \gamma=\sqrt{\frac{\Delta}{L}}\cdot\frac{1}{T^{\frac{1}{2}}}, h=2(\sqrt{L\Delta}T^{\frac{p+2}{6p-4}}+\|\nabla f(\bm{w}^{0})\|).$$ 
    This setting is identical to subcase 2) and the proof is similar.
    }  
    
	Combing the bounds \eqref{th1-c-b1}, \eqref{th1-c-b2} and \eqref{th1-c-b3}, we then complete {\color{black}the upper bound as  $L\Delta\geq 1$ as
     		\begin{align*}
			&\frac{1}{T}\sum_{t=1}^T\EE\|\nabla f(\bm{w}^{t})\|
\leq\frac{4(L\Delta)^{\frac{p-1}{3p-2}} \sigma^{\frac{p}{3p-2}}+4\sigma^{\frac{p}{3p-2}}}{T^{\frac{p-1}{3p-2}}}\\
&\quad +\frac{4 (L\Delta)^{\frac{p-1}{3p-2}}\sigma^{\frac{2p-2}{3p-2}}+16{\color{black} (L\Delta)^{\frac{p}{3p-2}}}\sigma^{\frac{p-1}{3p-2}}+16(L\Delta)^{\frac{6p-2p^2-2}{6p-4}}\sigma^{\frac{p^2}{6p-4}}+16(L\Delta)^{\frac{1}{4}}\sigma^{\frac{p}{8}}}{T^{\frac{p-1}{3p-2}}}\\
&\quad +\frac{16(L\Delta)^{\frac{p}{6p-4}}\|\nabla f(\bm{w}^{0})\|^{\frac{2 -p}{2}}\sigma^{\frac{3p^2-6p+4}{2(3p-2)}}}{T^{\frac{p}{6p-4}}}+\frac{2\|\nabla f(\bm{w}^0)\|\sigma^{\frac{4p-4}{3p-2}}+2\|\nabla f(\bm{w}^0)\|\sigma^{\frac{2p}{3p-2}}}{T^{\frac{2p-2}{3p-2}}}\\
&\quad +\frac{4\sqrt{L\Delta}+16\|\nabla f(\bm{w}^0)\|^{\frac{2 -p}{2}}+4}{T^{\frac{1}{2}}}+\frac{2\|\nabla f(\bm{w}^0)\|}{T}.
		\end{align*}

\textbf{Case II}: $L\Delta < 1$.  Following the same process in Case I, we can get
 		\begin{align*}
			&\frac{1}{T}\sum_{t=1}^T\EE\|\nabla f(\bm{w}^{t})\|
\leq\frac{L\Delta\sigma^{\frac{p}{3p-2}}+23\sigma^{\frac{p}{3p-2}}}{T^{\frac{p-1}{3p-2}}}+\frac{L\Delta\sigma^{\frac{2p-2}{3p-2}}+3\sigma^{\frac{2p-2}{3p-2}}+16\sigma^{\frac{p-1}{3p-2}}+16\sigma^{\frac{p}{8}}}{T^{\frac{p-1}{3p-2}}}\\
			&\quad+\frac{16\|\nabla f(\bm{w}^{0})\|^{\frac{2 -p}{2}}\sigma^{\frac{3p^2-6p+4}{2(3p-2)}}}{T^{\frac{p}{6p-4}}}+\frac{2\|\nabla f(\bm{w}^0)\|\sigma^{\frac{4p-4}{3p-2}}+2\|\nabla f(\bm{w}^0)\|\sigma^{\frac{2p}{3p-2}}}{T^{\frac{2p-2}{3p-2}}}\\
&\quad+\frac{L\Delta+16\|\nabla f(\bm{w}^0)\|^{\frac{2 -p}{2}}+7}{T^{\frac{1}{2}}}+\frac{2\|\nabla f(\bm{w}^0)\|}{T}\\
&\leq\frac{(L\Delta)^{\frac{p-1}{3p-2}}\sigma^{\frac{p}{3p-2}}+23\sigma^{\frac{p}{3p-2}}}{T^{\frac{p-1}{3p-2}}}+\frac{(L\Delta)^{\frac{p-1}{3p-2}}\sigma^{\frac{2p-2}{3p-2}}+3\sigma^{\frac{2p-2}{3p-2}}+16\sigma^{\frac{p-1}{3p-2}}+16\sigma^{\frac{p}{8}}}{T^{\frac{p-1}{3p-2}}}\\
			&\quad+\frac{16\|\nabla f(\bm{w}^{0})\|^{\frac{2 -p}{2}}\sigma^{\frac{3p^2-6p+4}{2(3p-2)}}}{T^{\frac{p}{6p-4}}}+\frac{2\|\nabla f(\bm{w}^0)\|\sigma^{\frac{4p-4}{3p-2}}+2\|\nabla f(\bm{w}^0)\|\sigma^{\frac{2p}{3p-2}}}{T^{\frac{2p-2}{3p-2}}}\\
&\quad+\frac{\sqrt{L\Delta}+16\|\nabla f(\bm{w}^0)\|^{\frac{2 -p}{2}}+7}{T^{\frac{1}{2}}}+\frac{2\|\nabla f(\bm{w}^0)\|}{T}
		\end{align*}
as $L\Delta<1$.

Combing the upper bounds in Case I and Case II together, we can get the final result.
}

	\section{Proof of Theorem \ref{th2-c}}
	This section employs   the following shorthand notation
	\begin{equation}\label{th2-c-temp1}
		\begin{aligned}
			&		\bm{\epsilon}^t:=\bm{m}^t-\nabla f(\bm{w}^t), \bm{\delta}^t:=\Ch_{h}(\nabla f(\bm{w}^{t};\xi^t))-\EE\Ch_{h}(\nabla f(\bm{w}^{t};\xi^t)),\\
			&\bm{c}^t:=\EE\Ch_{h}(\nabla f(\bm{w}^{t};\xi^t))-\nabla f(\bm{w}^t),\\
			&\hat{\bm{s}}^t:=\nabla f(\bm{w}^{t-1})-\nabla f(\bm{w}^{t})+\nabla f(\bm{w}^{t};\xi^t)-\nabla f(\bm{w}^{t-1};\xi^t).
		\end{aligned}
	\end{equation}
	Like the proof of NSGDC, we have
	\begin{align*}
		\bm{\epsilon}^t= \theta^t\bm{\epsilon}^{0}+\sum_{k=1}^{t}\theta^{t-k}\hat{\bm{s}}^k+(1-\theta)\sum_{k=1}^{t}\theta^{t-k}\bm{\delta}^k+(1-\theta)\sum_{k=1}^{t}\theta^{t-k}\bm{c}^k,
	\end{align*}
	which then yields
	\begin{align}\label{th2-c-temp2}
		\begin{aligned}
			\EE\|\bm{\epsilon}^t\|&\leq \EE\Big\|\sum_{k=1}^{t}\theta^{t-k}\hat{\bm{s}}^k\Big\|+(1-\theta)\sum_{k=1}^{t}\theta^{t-k}\EE\|\bm{c}^k\|+(1-\theta)\EE\Big\|\sum_{k=1}^{t}\theta^{t-k}\bm{\delta}^k\Big\|+\theta^t\|\bm{\epsilon}^{0}\|.
		\end{aligned}
	\end{align}
	The bound of $\EE\Big\|\sum_{k=1}^{t}\theta^{t-k}\bm{\delta}^k\Big\|$ can follow the same as that in the last proof of NSGDC. Noticing that $\EE\hat{\bm{s}}^k=\bm{0}$ and
	\begin{align*}
		\EE\|\hat{\bm{s}}^k\|^2&\leq 2\EE\|\nabla f(\bm{w}^{k-1})-\nabla f(\bm{w}^{k})\|^2+2\|\nabla f(\bm{w}^{k};\xi^k)-\nabla f(\bm{w}^{k-1};\xi^k)\|^2\\
		&\leq 4L^2\EE\|\bm{w}^{k}-\bm{w}^{k-1}\|^2\leq 4L^2\gamma^2,
	\end{align*}
	we can get
	\begin{align*}
		\EE\|\sum_{k=1}^{t}\theta^{t-k}\hat{\bm{s}}^k\|&\leq\sqrt{\EE\|\sum_{k=1}^{t}\theta^{t-k}\hat{\bm{s}}^k\|^2}=\sqrt{\sum_{k=1}^{t}\theta^{2t-2k}\EE\|\hat{\bm{s}}^k\|^2}\\
		&\leq\sqrt{\frac{4L^2\gamma^2}{1-\theta^2}}\leq\frac{2L\gamma}{\sqrt{1-\theta}},
	\end{align*}
	Based on \eqref{th2-c-temp2}, we then get
	\begin{align*} 
		\begin{aligned}
			\EE\|\bm{\epsilon}^t\|&\leq \frac{2L\gamma}{\sqrt{1-\theta}}  +2\sigma^{p}h^{-(p-1)}+4\sqrt{1-\theta}h^{\frac{2 -p}{2}}\sigma^{p/2}+\theta^t\|\bm{\epsilon}^{0}\|,
		\end{aligned}
	\end{align*}
	which further indicates
	\begin{align}\label{th2-c-temp3}
		\begin{aligned}
			\sum_{t=1}^T\EE\|\bm{\epsilon}^t\|/T&\leq \frac{2L\gamma}{\sqrt{1-\theta}}  +2\sigma^{p}h^{-(p-1)}+4\sqrt{1-\theta}h^{\frac{2 -p}{2}}\sigma^{p/2} +\frac{\|\nabla f(\bm{w}^0)\|}{(1-\theta)T}.
		\end{aligned}
	\end{align}
	Note that \eqref{conver-core} still holds,  it then follows
	\begin{align}\label{th2-c-bound}
		\begin{aligned}
			&\frac{1}{T}\sum_{t=1}^T\EE\|\nabla f(\bm{w}^{t})\|\\
			&\leq \frac{f(\bm{w}^{1})-\min f}{\gamma T}+
			L\gamma/2+\frac{4L\gamma}{\sqrt{1-\theta}} +\frac{2\|\nabla f(\bm{w}^0)\|}{(1-\theta)T}\\
            &\qquad+4\sigma^{p}h^{-(p-1)}+8\sqrt{1-\theta}h^{\frac{2 -p}{2}}\sigma^{p/2}\\
			&\leq   \frac{f(\bm{w}^{1})-\min f}{\gamma T}+\frac{5L\gamma}{\sqrt{1-\theta}} +\frac{2\|\nabla f(\bm{w}^0)\|}{(1-\theta)T}+4\sigma^{p}h^{-(p-1)}+8\sqrt{1-\theta}h^{\frac{2 -p}{2}}\sigma^{p/2}.
		\end{aligned}
	\end{align}
	\medskip

	{\color{black}
	\noindent	1). If $\sigma\leq 1$ and  ${\color{black}\sigma^{\frac{2p-2}{2p-1}}}T^{\frac{p}{2p-1}}\geq 1$, we have
	$$1-\theta=\frac{1}{{\color{black}\sigma^{\frac{2p-2}{2p-1}}}T^{\frac{p}{2p-1}}},\gamma=\frac{1}{L{\color{black}\sigma^{\frac{p-1}{4p-2}}}T^{\frac{5p-2}{8p-4}}}, h=\frac{2T^{\frac{1}{2p-1}}}{\sigma^{\frac{p-1}{4p-2}}}+2\|\nabla f(\bm{w}^{0})\|.$$
The upper bound of \eqref{th2-c-bound} can be reformulated as
	\begin{align}\label{th2-c-b1}
		\begin{aligned}
			\frac{1}{T}\sum_{t=1}^T\EE\|\nabla f(\bm{w}^{t})\|
			&\leq   \frac{[L(f(\bm{w}^{1})-\min f)+5]{\color{black}\sigma^{\frac{p-1}{4p-2}}}}{ T^{\frac{3p-2}{8p-4}}}+\frac{2\|\nabla f(\bm{w}^0)\|{\color{black}\sigma^{\frac{2p-2}{2p-1}}}}{T^{\frac{p-1}{2p-1}}}+\frac{4\sigma^{p+\frac{(p-1)^2}{4p-2}}}{T^{\frac{p-1}{2p-1}}}\\
&\qquad+\frac{16\sigma^{\frac{5p^2-9p+6}{8p-4}}}{T^{\frac{5p-1}{8p-4}}}+\frac{16\|\nabla f(\bm{w}^{0})\|^{\frac{2-p}{2}}\sigma^{\frac{2p^2-5p+2}{4p-2}}}{T^{\frac{p}{4p-2}}}\\
			&\leq   \frac{[L(f(\bm{w}^{1})-\min f)+5]{\color{black}\sigma^{\frac{p-1}{4p-2}}}}{ T^{\frac{3p-2}{8p-4}}}+\frac{2\|\nabla f(\bm{w}^0)\|{\color{black}\sigma^{\frac{2p-2}{2p-1}}}}{T^{\frac{p-1}{2p-1}}}+\frac{4\sigma^{\frac{p}{2p-1}}}{T^{\frac{p-1}{2p-1}}}\\
&\qquad+\frac{16\sigma^{\frac{5p^2-9p+6}{8p-4}}}{T^{\frac{5p-1}{8p-4}}}+\frac{16\|\nabla f(\bm{w}^{0})\|^{\frac{2-p}{2}}\sigma^{\frac{2p^2-5p+2}{4p-2}}}{T^{\frac{p}{4p-2}}},
		\end{aligned}
	\end{align}
where we used $p+\frac{(p-1)^2}{4p-2}\geq \frac{p}{2p-1}$ as $p\in (1,2]$ and $0\leq \sigma\leq 1$.
	}

{\color{black}
	\noindent	2). If  $\sigma\leq 1$ and  $\sigma^{\frac{2p-2}{2p-1}}T^{\frac{p}{2p-1}}\leq 1$, we have
	$$1-\theta=1,\gamma=\frac{1}{L\sqrt{T}}, h=2T^{\frac{p+4}{8p-4}}+2\|\nabla f(\bm{w}^{0})\|.$$
	Based on these hyperparameters, 
	\begin{align}\label{th2-c-b2}
		\begin{aligned}
			&\frac{1}{T}\sum_{t=1}^T\EE\|\nabla f(\bm{w}^{t})\|\leq   \frac{L(f(\bm{w}^{1})-\min f)+5}{ T^{\frac{1}{2}}}+\frac{2\|\nabla f(\bm{w}^0)\|}{T}+\frac{4}{T^{\frac{p^2}{2p-2}+\frac{(p-1)(p+4)}{8p-4}}}\\
			&\qquad\qquad+\frac{16\sigma^{\frac{p}{16}}}{T^{\frac{7p^2}{32p-32}-\frac{(2-p)(p+4)}{16p-8}}}+\frac{16\|\nabla f(\bm{w}^{0})\|^{\frac{2-p}{2}}}{T^{\frac{p^2}{4p-4}}}\\
			&\quad\leq \frac{\sigma^{\frac{p}{16}}}{T^{\frac{p-1}{2p-1}}}+\frac{L(f(\bm{w}^{1})-\min f)+16\|\nabla f(\bm{w}^{0})\|^{\frac{2-p}{2}}+9}{ T^{\frac{1}{2}}}+\frac{2\|\nabla f(\bm{w}^0)\|}{T},
		\end{aligned}
	\end{align}
	because $\frac{p^2}{2p-2}+\frac{(p-1)(p+4)}{8p-4}\geq\frac{1}{2}$, $\frac{p^2}{4p-4}\geq\frac{1}{2}$, and $\frac{7p^2}{32p-32}-\frac{(2-p)(p+4)}{16p-8}\geq\frac{p-1}{2p-1}$ as $1< p\leq 2$.
}

	{\color{black}
	\noindent	3). If $\sigma\geq 1$, we have ${\color{black}\sigma^{\frac{p}{2p-1}}}T^{\frac{p}{2p-1}}\geq 1$, it then follows
	$$1-\theta=\frac{1}{{\color{black}\sigma^{\frac{p}{2p-1}}}T^{\frac{p}{2p-1}}},\gamma=\frac{1}{L{\color{black}\sigma^{\frac{p}{8p-4}}}T^{\frac{5p-2}{8p-4}}}, h=2T^{\frac{1}{2p-1}}\sigma^{\frac{2p}{2p-1}}+2\|\nabla f(\bm{w}^{0})\|.$$
	We can rebound \eqref{th2-c-bound} as
	\begin{align}\label{th2-c-b3}
		\begin{aligned}
			\frac{1}{T}\sum_{t=1}^T\EE\|\nabla f(\bm{w}^{t})\|
			&\leq   \frac{[L(f(\bm{w}^{1})-\min f)+5]{\color{black}\sigma^{\frac{p}{8p-4}}}}{ T^{\frac{3p-2}{8p-4}}}+\frac{2\|\nabla f(\bm{w}^0)\|{\color{black}\sigma^{\frac{p}{2p-1}}}}{T^{\frac{p-1}{2p-1}}}+\frac{4\sigma^{\frac{p}{2p-1}}}{T^{\frac{p-1}{2p-1}}}\\
&\qquad+\frac{16\sigma^{\frac{p}{2p-1}}}{T^{\frac{p-1}{2p-1}}}+\frac{16\|\nabla f(\bm{w}^{0})\|^{\frac{2-p}{2}}\sigma^{\frac{p^2-p}{2p-1}}}{T^{\frac{p}{4p-2}}}.
		\end{aligned}
	\end{align}
	}

		{\color{black}
	The proof is completed by combining the bounds from equations \eqref{th2-c-b1}, \eqref{th2-c-b2} and \eqref{th2-c-b2}, along with the condition that $\bm{w}^0 = \bm{w}^1$.
	}
	
	\section{Proof of Theorem \ref{th3-c}}
	This section recruits the following notation
	\begin{align*}
		&\bm{\epsilon}^t:=\bm{m}^t-\nabla f(\bm{w}^t), \bm{\delta}^t:=\Ch_{h}(\bm{g}^t)-\EE\Ch_{h}(\bm{g}^t),\\
		&\bm{c}^t:=\EE\Ch_{h}(\bm{g}^t)-\nabla f(\bm{v}^t),~{\bf D}(\bm{x},\bm{y}):=\nabla f(\bm{x})-\nabla f(\bm{y})-[\nabla^2 f(\bm{x})](\bm{x}-\bm{y}).
	\end{align*}
	If  Assumption \ref{ass3} holds, for any $\bm{x},\bm{y}\in\RR$, the two-point operator satisfies
	\begin{align*}
		\|{\bf D}(\bm{x},\bm{y})\|&=\|\nabla f(\bm{x})-\nabla f(\bm{y})-[\nabla^2 f(\bm{x})](\bm{x}-\bm{y})\|\\
		&=\Big\|\int_{s=0}^{1}[\nabla^2 f(\bm{y}+(\bm{x}-\bm{y})s)-\nabla^2 f(\bm{x})](\bm{x}-\bm{y}) ds \Big\|\\
		&\leq \int_{s=0}^{1}\|\nabla^2 f(\bm{y}+(\bm{x}-\bm{y})s)-\nabla^2 f(\bm{x})\|_{\textrm{op}}\|\bm{x}-\bm{y}\|ds\\
		&\leq H\int_{s=0}^{1}\|\bm{x}-\bm{y}\|^2ds=\frac{H}{2}\|\bm{x}-\bm{y}\|^2.
	\end{align*}
	The scheme of the algorithm gives us
	\begin{align*}
		\bm{m}^{t}&=\theta\bm{m}^{t-1}+(1-\theta)\Ch_{h}(\bm{g}^t)=\theta(\bm{\epsilon}^{t-1}+\nabla f(\bm{w}^{t-1}))+(1-\theta)(\bm{\delta}^t+\bm{c}^t+\nabla f(\bm{v}^t))\\
		&=\theta(\bm{\epsilon}^{t-1}+\nabla f(\bm{w}^{t})+[\nabla^2 f(\bm{w}^{t})](\bm{w}^{t-1}-\bm{w}^{t})+{ \bf D}(\bm{w}^{t-1},\bm{w}^{t}))\\
		&\quad\qquad+(1-\theta)(\bm{\delta}^t+\bm{c}^t+\nabla f(\bm{w}^{t})+[\nabla^2 f(\bm{w}^{t})](\bm{v}^{t}-\bm{w}^{t})+{\bf D}(\bm{v}^{t},\bm{w}^{t}))\\
		&=\theta(\bm{\epsilon}^{t-1}+\nabla f(\bm{x}^{t})+{\bf D}(\bm{w}^{t-1},\bm{w}^{t}))+(1-\theta)(\bm{\delta}^t+\bm{c}^t+\nabla f(\bm{w}^{t})+{\bf D}(\bm{v}^{t},\bm{w}^{t})),
	\end{align*}
	where we use the fact $\bm{v}^{t}-\bm{w}^{t}=\frac{\theta}{1-\theta}(\bm{w}^{t}-\bm{w}^{t-1})$.
	Based on the relation above, we further get
	$$	\bm{\epsilon}^t=\bm{m}^{t}-\nabla f(\bm{x}^t)=\theta\bm{\epsilon}^{t-1}+\theta{\bf H}(\bm{w}^{t-1},\bm{w}^{t})+(1-\theta){\bf H}(\bm{v}^{t},\bm{w}^{t})+(1-\theta)(\bm{\delta}^t+\bm{c}^t).$$
	Recursion of iteration above from $0$ to $t$, we have
	\begin{align*}
		\bm{\epsilon}^t=  \theta\sum_{i=1}^{t-1}\theta^{t-i}{\bf H}(\bm{w}^{i-1},\bm{w}^{i})+(1-\theta)\sum_{i=1}^{t-1}\theta^{t-i}{\bf H}(\bm{v}^{i},\bm{w}^{i})+(1-\theta)\sum_{i=1}^{t-1}\theta^{t-i}(\bm{\delta}^i+\bm{c}^i)+\theta^t\bm{\epsilon}^{0}.
	\end{align*}
	Considering the norms and expectations of both sides of the equation above,
	\begin{align*}
		\EE\| \bm{\epsilon}^t\|&\leq   \theta\sum_{i=1}^{t-1}\theta^{t-i}\EE\|{\bf H}(\bm{w}^{i-1},\bm{w}^{i})\|+(1-\theta)\sum_{i=1}^{t-1}\theta^{t-i}\EE\|{\bf H}(\bm{v}^{i},\bm{w}^{i})\|\\
		&\qquad+\EE\|(1-\theta)\sum_{i=1}^{t-1}\theta^{t-i}(\bm{\delta}^i+\bm{c}^i)\|+\theta^t\|\bm{\epsilon}^{0}\|\\ &\leq\frac{H}{2}\theta\sum_{i=1}^{t-1}\theta^{t-i}\EE\|\bm{w}^{i-1}-\bm{w}^{i}\|^2+\frac{H}{2}(1-\theta)\sum_{i=1}^{t-1}\theta^{t-i}\EE\|\bm{v}^{i}-\bm{w}^{i}\|^2\\
		&\qquad+\EE\|(1-\theta)\sum_{i=1}^{t-1}\theta^{t-i}(\bm{\delta}^i+\bm{c}^i)\|+\theta^t\|\bm{\epsilon}^{0}\|\\ 
		&\leq\frac{H}{2}\frac{\theta}{1-\theta}\sum_{i=1}^{t-1}\theta^{t-i}\EE\|\bm{w}^{i-1}-\bm{w}^{i}\|^2+\EE\|(1-\theta)\sum_{i=1}^{t-1}\theta^{t-i}(\bm{\delta}^i+\bm{c}^i)\|+\theta^t\|\bm{\epsilon}^{0}\|\\
		&\leq \frac{H}{2}\frac{\theta}{(1-\theta)^2}\gamma^2+\EE\|(1-\theta)\sum_{i=1}^{t-1}\theta^{t-i}\bm{\delta}^i\|+\EE\|(1-\theta)\sum_{i=1}^{t-1}\theta^{t-i}\bm{c}^i\|+\theta^t\|\bm{\epsilon}^{0}\|,
	\end{align*}
	where we used the   gradient normalization scheme
	$$\EE\|\bm{w}^{i-1}-\bm{w}^{i}\|^2\leq \gamma^2.$$
	The bounds for $\EE\|(1-\theta)\sum_{i=1}^{t-1}\theta^{t-i}\bm{\delta}^i\|$ and $\EE\|(1-\theta)\sum_{i=1}^{t-1}\theta^{t-i}\bm{c}^i\|$ are analogous to that presented in Theorem \ref{th1-c}. We   then get
	\begin{align*}
		\EE\| \bm{\epsilon}^t\|\leq \frac{H}{2}\frac{\theta}{(1-\theta)^2}\gamma^2+ 2\sigma^{p}h^{-(p-1)}+4\sqrt{1-\theta}h^{\frac{2 -p}{2}}\sigma^{p/2}+\theta^t\|\bm{\epsilon}^{0}\|.
	\end{align*}
	Summing the inequality from $t=1$ to $T$,
	\begin{align*}
		\sum_{t=1}^T\EE\| \bm{\epsilon}^t\|/T\leq \frac{H}{2}\frac{\theta}{(1-\theta)^2}\gamma^2+ 2\sigma^{p}h^{-(p-1)}+4\sqrt{1-\theta}h^{\frac{2 -p}{2}}\sigma^{p/2}+\frac{\|\bm{\epsilon}^{0}\|}{(1-\theta)T}.
	\end{align*}
	Employing Lemma \ref{lemma1} with  $\bm{w}^{\dag}\rightarrow \bm{w}^{t}$ and $\bm{m}\rightarrow \bm{w}^{t}$,
	\begin{align}\label{th3-reuse}
		f(\bm{w}^{t+1})-f(\bm{w}^{t})\leq -\gamma \|\nabla f(\bm{w}^{t})\|+2\gamma\|\bm{\epsilon}^t\|+\frac{L\gamma^2}{2}.
	\end{align}
	Summing the inequality from $t=1$ to $T$,
	\begin{align}\label{th3-c-temp}
		\begin{aligned}
			\frac{1}{T}\sum_{t=1}^T\EE\|\nabla f(\bm{w}^{t})\|
			&\leq \frac{f(\bm{w}^{1})-\min f}{\gamma T}+ \frac{H\theta}{(1-\theta)^2}\gamma^2 \\
			&+ 4\sigma^{p}h^{-(p-1)}+8\sqrt{1-\theta}h^{\frac{2 -p}{2}}\sigma^{p/2}+\frac{2\|\bm{\epsilon}^{0}\|}{(1-\theta)T}+\frac{L\gamma }{2}.
		\end{aligned}
	\end{align}
 
{\color{black}
	\noindent 1) 
	As $0\leq \sigma\leq 1$ and $\sigma^{3/7} T^{\frac{2p}{4p-1}}>1$, and then
	$$1-\theta=\frac{1}{\sigma^{\frac{3}{7}} T^{{\frac{2p}{4p-1}}}},~\gamma=\frac{1}{L\sigma^{\frac{2}{7}}T^{\frac{2p+1}{4p-1}}}, h=2\Big(\frac{T^{\frac{2}{4p-1}}}{\sigma^{\frac{2}{7}}}+\|\nabla f(\bm{w}^{0})\|\Big).$$
	We then have the following rate as
	\begin{align}\label{th3-c-b1-pre}
		\begin{aligned}
			&\frac{1}{T}\sum_{t=1}^T\EE\|\nabla f(\bm{w}^{t})\|
			\leq \frac{L\sigma^{\frac{2}{7}}(f(\bm{w}^{1})-\min f)}{T^{\frac{2p-2}{4p-1}}} +\frac{H\sigma^{\frac{2}{7}}}{L^2T^{\frac{2}{4p-1}}}  \\
			&\quad+ \frac{4\sigma^{p+\frac{2(p-1)}{7}}}{T^{\frac{2p-2}{4p-1}}} +16\sqrt{1-\theta}\Big(\frac{T^{\frac{2}{4p-1}}}{\sigma^{\frac{2}{7}}}+\|\nabla f(\bm{w}^{0})\|\Big)^{\frac{2 -p}{2}}\sigma^{p/2}+\frac{2\|\bm{\epsilon}^{0}\|\sigma^{\frac{3}{7}}}{T^{\frac{2p-1}{4p-1}}}+\frac{1 }{2\sigma^{\frac{2}{7}}T^{\frac{2p+1}{4p-1}}}\\
			&\leq \frac{L\sigma^{\frac{2}{7}}(f(\bm{w}^{1})-\min f)}{T^{\frac{2p-2}{4p-1}}} +\frac{H\sigma^{\frac{2}{7}}}{L^2T^{\frac{2}{4p-1}}} + \frac{4\sigma^{p}}{T^{\frac{2p-2}{4p-1}}}+\frac{2\|\bm{\epsilon}^{0}\|\sigma^{\frac{3}{7}}}{T^{\frac{2p-1}{4p-1}}}+\frac{\sigma^{\frac{9-4p}{14p}}}{2T^{\frac{2p-2}{4p-1}}}\\
			&\quad +\frac{16\sigma^{\frac{9p-7}{14}}}{T^{\frac{2p-2}{4p-1}}}+\frac{16\|\nabla f(\bm{w}^{0})\|^{\frac{2 -p}{2}}\sigma^{\frac{7p-3}{14}}}{T^{\frac{p}{4p-1}}}.
		\end{aligned}
	\end{align}
	Due to  $0\leq \sigma\leq 1$ and $1<p\leq 2$, we have
$$\sigma^{p+\frac{2(p-1)}{7}}\leq \sigma^{\frac{2}{7}}, \sigma^{\frac{9p-7}{14}}\leq \sigma^{\frac{1}{7}},\sigma^{\frac{9-4p}{14p}}\leq\sigma^{\frac{1}{28}}.$$
Thus, we are then led to
\begin{align}\label{th3-c-b1}
		\begin{aligned}
			\frac{1}{T}\sum_{t=1}^T\EE\|\nabla f(\bm{w}^{t})\|
			&\leq \frac{L\sigma^{\frac{2}{7}}(f(\bm{w}^{1})-\min f)}{T^{\frac{2p-2}{4p-1}}} +\frac{H\sigma^{\frac{2}{7}}}{L^2T^{\frac{2}{4p-1}}} + \frac{20\sigma^{\frac{2}{7}}}{T^{\frac{2p-2}{4p-1}}}\\
			&\quad+\frac{2\|\bm{\epsilon}^{0}\|\sigma^{\frac{3}{7}}}{T^{\frac{2p-1}{4p-1}}} +\frac{16\|\nabla f(\bm{w}^{0})\|^{\frac{2 -p}{2}}\sigma^{\frac{2}{7}}}{T^{\frac{p}{4p-1}}}+\frac{\sigma^{\frac{1}{28}}}{2T^{\frac{2p+1}{4p-1}}}.
		\end{aligned}
	\end{align}

	\noindent 2) 
	When $0\leq \sigma\leq 1$ and  $\sigma^{3/7} T^{\frac{2p}{4p-1}}\leq 1$, we have
$$1-\theta=1,~\gamma=\frac{1}{LT^{\frac{1}{2}}}, h=2(T^{\frac{7}{8p-2}}+\|\nabla f(\bm{w}^{0})\|).$$
	Recalling \eqref{th3-c-temp}, we then get
	\begin{align}\label{th3-c-b2}
		\begin{aligned}
			&\frac{1}{T}\sum_{t=1}^T\EE\|\nabla f(\bm{w}^{t})\|
			\leq \frac{L(f(\bm{w}^{1})-\min f)+1}{T^{1/2}}+ 4T^{-\frac{4p^2+7p-7}{8p-2}} \\
			&\qquad+\frac{16\sigma^{\frac{p}{16}}}{T^{\frac{7(p+2)(p-1)}{16p-4}}}+16\|\nabla f(\bm{w}^{0})\|^{\frac{2 -p}{2}}T^{-\frac{2p^2}{4p-1}}+\frac{2\|\bm{\epsilon}^{0}\|}{T}\\
			&\leq \frac{L(f(\bm{w}^{1})-\min f)+5+16\|\nabla f(\bm{w}^{0})\|^{\frac{2 -p}{2}}}{T^{1/2}}+\frac{16\sigma^{\frac{p}{16}}}{T^{\frac{2p-2}{4p-1}}}+\frac{2\|\bm{\epsilon}^{0}\|}{T}.
		\end{aligned}
	\end{align}

	\noindent 3) 
	As $\sigma\geq 1$, we have $\sigma^{6/7} T^{\frac{2p}{4p-1}}> 1$ due to $T> 1$, and then
	$$1-\theta=\frac{1}{\sigma^{\frac{6}{7}} T^{{\frac{2p}{4p-1}}}},~\gamma=\frac{1}{L\sigma^{\frac{4}{7}}T^{\frac{2p+1}{4p-1}}}, h=2\Big(T^{\frac{2}{4p-1}}\cdot\sigma^{\frac{7p-4}{7p-7}}+\|\nabla f(\bm{w}^{0})\|\Big).$$
	We then have the following rate as
	\begin{align}\label{th3-c-b3-pre}
		\begin{aligned}
			&\frac{1}{T}\sum_{t=1}^T\EE\|\nabla f(\bm{w}^{t})\|
			\leq \frac{L\sigma^{\frac{4}{7}}(f(\bm{w}^{1})-\min f)}{T^{\frac{2p-2}{4p-1}}} +\frac{H\sigma^{\frac{4}{7}}}{L^2T^{\frac{2}{4p-1}}}  \\
			&\quad+ \frac{4\sigma^{\frac{4}{7}}}{T^{\frac{2p-2}{4p-1}}} +16\sqrt{1-\theta}\Big(T^{\frac{2}{4p-1}}\cdot\sigma^{\frac{7p-4}{7p-7}}+\|\nabla f(\bm{w}^{0})\|\Big)^{\frac{2 -p}{2}}\sigma^{p/2}+\frac{2\|\bm{\epsilon}^{0}\|\sigma^{\frac{6}{7}}}{T^{\frac{2p-1}{4p-1}}}+\frac{1 }{2\sigma^{\frac{4}{7}}T^{\frac{2p+1}{4p-1}}}\\
			&\leq \frac{L\sigma^{\frac{4}{7}}(f(\bm{w}^{1})-\min f)}{T^{\frac{2p-2}{4p-1}}} +\frac{H\sigma^{\frac{4}{7}}}{L^2T^{\frac{2}{4p-1}}} + \frac{4\sigma^{\frac{4}{7}}}{T^{\frac{2p-2}{4p-1}}}+\frac{2\|\bm{\epsilon}^{0}\|\sigma^{\frac{6}{7}}}{T^{\frac{2p-1}{4p-1}}}+\frac{1 }{2\sigma^{\frac{4}{7}}T^{\frac{2p+1}{4p-1}}}\\
			&\quad +\frac{16\sigma^{\frac{p}{2}-\frac{3}{7}-\frac{(2-p)(7p-4)}{14p-14}}}{T^{\frac{2p-2}{4p-1}}}+\frac{16\|\nabla f(\bm{w}^{0})\|^{\frac{2 -p}{2}}\sigma^{\frac{p}{2}-\frac{3}{7}}}{T^{\frac{p}{4p-1}}}.
		\end{aligned}
	\end{align}
	Using the fact $\sigma\geq 1$ and $1< p\leq 2$, we have
$$\frac{1}{\sigma^{\frac{4}{7}}}\leq \sigma^{\frac{4}{7}}, \sigma^{\frac{p}{2}-\frac{3}{7}-\frac{(2-p)(7p-4)}{14p-14}}\leq \sigma^{\frac{4}{7}}, \sigma^{\frac{p}{2}-\frac{3}{7}}\leq \sigma^{\frac{4}{7}}.$$
Thus, we are then led to
\begin{align}\label{th3-c-b3}
		\begin{aligned}
			\frac{1}{T}\sum_{t=1}^T\EE\|\nabla f(\bm{w}^{t})\|
			&\leq \frac{L\sigma^{\frac{4}{7}}(f(\bm{w}^{1})-\min f)}{T^{\frac{2p-2}{4p-1}}} +\frac{H\sigma^{\frac{4}{7}}}{L^2T^{\frac{2}{4p-1}}} + \frac{20\sigma^{\frac{4}{7}}}{T^{\frac{2p-2}{4p-1}}}\\
			&\quad+\frac{2\|\bm{\epsilon}^{0}\|\sigma^{\frac{6}{7}}}{T^{\frac{2p-1}{4p-1}}}+\frac{\sigma^{\frac{4}{7}}}{2T^{\frac{2p+1}{4p-1}}} +\frac{16\|\nabla f(\bm{w}^{0})\|^{\frac{2 -p}{2}}\sigma^{\frac{4}{7}}}{T^{\frac{p}{4p-1}}}.
		\end{aligned}
	\end{align}

	We then proved the result by combining \eqref{th3-c-b1}, \eqref{th3-c-b2} and \eqref{th3-c-b3}.
}
	\section{Accelerated Normalized Stochastic Gradient Descent under Second-Order Smoothness }\label{NSGD2}
	\begin{algorithm}[H]
		\caption{Accelerated Normalized Stochastic Gradient Descent under Second-Order Smoothness (A-NSGD)}\label{alg3}
		\begin{algorithmic}[1]
			\REQUIRE   parameters $\gamma>0$,  $0\leq \theta<1$, $\zeta\geq 0$\\
			\textbf{Initialization}: $\bm{w}^0=\bm{w}^1$, $\bm{m}^0=\bm{0}$\\
			\textbf{for}~$t=1,2,\ldots$ \\
			~~\textbf{step 1} $\bm{v}^t=\bm{w}^{t}+\zeta(\bm{w}^{t}-\bm{w}^{t-1})$\\
			~~\textbf{step 2}: Sample the data  $\xi^t\sim\mathcal{D}$ and $\bm{m}^t=\theta\bm{m}^{t-1}+(1-\theta)\nabla f(\bm{v}^t;\xi^t)$\\
			~~\textbf{step 3}: $ \bm{w}^{t+1}= \bm{w}^{t}-\gamma \frac{\bm{m}^t}{\|\bm{m}^t\|}$ \\
			\textbf{end for}\\
		\end{algorithmic}
	\end{algorithm}
	The convergence result of nonconvex accelerated stochastic algorithm under heavy-tailed noise is presented as follows.
	\begin{proposition}\label{th3}
		Let $(\bm{w}^t)_{t\geq0}$ be generated by Algorithm \ref{alg3}. Suppose Assumption \ref{ass2}--\ref{ass3} hold.
		If 		$1-\theta=\min\Big\{\frac{1}{\sigma^\frac{1}{2} T^{\frac{2p}{4p-1}}},1\Big\}$, $\gamma=\frac{(1-\theta)^{\frac{3}{4p}}}{L\sqrt{T}}$, $\zeta=\frac{\theta}{1-\theta}$,
		then it holds that
		\begin{align*}
			&\frac{1}{T}\sum_{t=1}^T\EE\|\nabla f(\bm{w}^{t})\|\leq \frac{L\sigma^{\frac{3}{8p}}(f(\bm{w}^{1})-\min f)+\frac{H}{L^2}\sigma^{\frac{4p-3}{4p}}+4\Big(\frac{B}{L}+1\Big)^{\frac{2 -p}{2}}(\sigma^{\frac{1}{5}}+\sigma^{\frac{1}{2}})+\sigma^{\frac{3}{8p}}}{T^{\frac{2p-2}{4p-1}}} \\
			&\qquad+\frac{\|\nabla f(\bm{w}^0)\|\sigma^{\frac{1}{2}}}{T^{\frac{2p-1}{4p-1}}}+\frac{L(f(\bm{w}^{0})-\min f)+1+4\Big(\frac{B}{L}+1\Big)^{\frac{2 -p}{2}}}{ \sqrt{T}}+\frac{2\|\nabla f(\bm{w}^0)\|}{T}.
		\end{align*}
	\end{proposition}
	{\bf	Proof of Proposition \ref{th3}}:
	This proof employs the following notation
	\begin{align*}
		\bm{g}^t:=\nabla f(\bm{v}^t;\xi^t),\bm{\delta}^t:=\nabla f(\bm{v}^t;\xi^t)-\nabla f(\bm{v}^t),\\
		~\bm{\epsilon}^t:=\bm{m}^t-\nabla f(\bm{w}^t),~{\bf D}(\bm{x},\bm{y}):=\nabla f(\bm{x})-\nabla f(\bm{y})-[\nabla^2 f(\bm{x})](\bm{x}-\bm{y}).
	\end{align*}
	Similar to the proof of Theorem \ref{th3-c}, we can derive
	\begin{align*}
		\EE\| \bm{\epsilon}^t\|\leq \frac{H}{2}\frac{\theta}{(1-\theta)^2}\gamma^2+\EE\|(1-\theta)\sum_{i=1}^{t-1}\theta^{t-i}\bm{\delta}^i\|+\theta^t\|\bm{\epsilon}^{0}\|.
	\end{align*}
	The bound of $\EE\|(1-\theta)\sum_{i=1}^{t-1}\theta^{t-i}\bm{\delta}^i\|$ follows the same of that in Theorem \ref{th1}.
	Thus, we are led to the following bound
	\begin{align*}
		\EE\| \bm{\epsilon}^t\|\leq \frac{H}{2}\frac{\theta}{(1-\theta)^2}\gamma^2+2\sqrt{1-\theta}(B+L\gamma T)^{\frac{2 -p}{2}}\sigma^{p/2}+\theta^t\|\bm{\epsilon}^{0}\|,
	\end{align*}
	which yields
	\begin{align*}
		\frac{1}{T}\sum_{t=1}^T \EE\| \bm{\epsilon}^t\|\leq \frac{H}{2}\frac{\theta}{(1-\theta)^2}\gamma^2+2\sqrt{1-\theta}(B+L\gamma T)^{\frac{2 -p}{2}}\sigma^{p/2}+\frac{\|\bm{\epsilon}^{0}\|}{(1-\theta)T}.
	\end{align*}
	Noticing \eqref{th3-reuse} still holds,  aggregating the inequality over the range from $t=1$ to $T$,
	\begin{align*}
		&\frac{1}{T}\sum_{t=1}^T\EE\|\nabla f(\bm{w}^{t})\|\\
		&\leq \frac{f(\bm{w}^{1})-\min f}{\gamma T} + \frac{H\theta}{(1-\theta)^2}\gamma^2+4\sqrt{1-\theta}(B+L\gamma T)^{\frac{2 -p}{2}}\sigma^{p/2}+\frac{L\gamma }{2}+\frac{2\|\bm{\epsilon}^{0}\|}{(1-\theta)T}\\
		&\leq \frac{f(\bm{w}^{1})-\min f}{\gamma T} + \frac{H\theta}{(1-\theta)^2}\gamma^2+4(B+L)^{\frac{2 -p}{2}}\sqrt{1-\theta}(\gamma T)^{\frac{2 -p}{2}}\sigma^{p/2}+\frac{L\gamma }{2}+\frac{2\|\bm{\epsilon}^{0}\|}{(1-\theta)T}.
	\end{align*}
	
	\medskip
	
	\noindent 1) If $\sigma^{1/2} T^{\frac{2p}{4p-1}}\geq 1$, we have
	$$1-\theta=\frac{1}{\sigma^{\frac{1}{2}} T^{{\frac{2p}{4p-1}}}},~\gamma=\frac{1}{L\sigma^{\frac{3}{8p}}T^{\frac{2p+1}{4p-1}}}.$$
	The bound of rate then reduces to
	\begin{align}\label{th3-b1}
		\begin{aligned}
			&\frac{1}{T}\sum_{t=1}^T\EE\|\nabla f(\bm{w}^{t})\|\\
			&\leq \frac{L\sigma^{\frac{3}{8p}}(f(\bm{w}^{1})-\min f)}{T^{\frac{2p-2}{4p-1}}} +\frac{H\sigma^{\frac{4p-3}{4p}}}{L^2T^{\frac{2}{4p-1}}} +4\Big(\frac{B}{L}+1\Big)^{\frac{2 -p}{2}}\frac{\sigma^{\frac{8p^2-11p+6}{16p}}}{T^{\frac{p^2-2p+2}{4p-1}}} +\frac{1 }{2\sigma^{\frac{3}{8p}}T^{\frac{2p+1}{4p-1}}}+\frac{\|\bm{\epsilon}^{0}\|\sigma^{\frac{1}{2}}}{T^{\frac{2p-1}{4p-1}}}\\
			&\leq \frac{L\sigma^{\frac{3}{8p}}(f(\bm{w}^{1})-\min f)}{T^{\frac{2p-2}{4p-1}}} +\frac{H\sigma^{\frac{4p-3}{4p}}}{L^2T^{\frac{2p-2}{4p-1}}} +4\Big(\frac{B}{L}+1\Big)^{\frac{2 -p}{2}}\frac{(\sigma^{1/5}+\sigma^{1/2})}{T^{\frac{2p-2}{4p-1}}} +\frac{\sigma^{\frac{3}{8p}} }{2T^{\frac{2p-2}{4p-1}}}+\frac{\|\bm{\epsilon}^{0}\|\sigma^{\frac{1}{2}}}{T^{\frac{2p-1}{4p-1}}},
		\end{aligned}
	\end{align}
	where we used $\sigma^{\frac{8p^2-11p+6}{16p}} \leq\sigma^{1/5}+\sigma^{1/2}$ as $p\in (1,2]$.
	
	\medskip
	
	\noindent 2) If $\sigma^{1/2} T^{\frac{2p}{4p-1}}\leq 1$, we have
	$$1-\theta=1,~\gamma=\frac{1}{L\sqrt{T}}.$$
	Further, we are led to
	\begin{align}\label{th3-b2}
		\begin{aligned}
			\frac{1}{T}\sum_{t=1}^T\EE\|\nabla f(\bm{w}^{t})\|
			&\leq \frac{L(f(\bm{w}^{1})-\min f)+1}{ \sqrt{T}} +4\Big(\frac{B}{L}+1\Big)^{\frac{2 -p}{2}}T^{\frac{2 -p}{4}}\sigma^{p/2}+\frac{2\|\bm{\epsilon}^{0}\|}{T}\\	
			&\leq \frac{L(f(\bm{w}^{1})-\min f)+1}{ \sqrt{T}} +4\Big(\frac{B}{L}+1\Big)^{\frac{2 -p}{2}}T^{\frac{2 -p}{4}-\frac{p(8-2p)}{11-2p}}+\frac{2\|\bm{\epsilon}^{0}\|}{T}\\
			&\leq \frac{L(f(\bm{w}^{0})-\min f)+1}{ \sqrt{T}} +\frac{4\Big(\frac{B}{L}+1\Big)^{\frac{2 -p}{2}}}{\sqrt{T}}+\frac{2\|\bm{\epsilon}^{0}\|}{T},
		\end{aligned}
	\end{align}
	where we used that
	$$\frac{p(8-2p)}{11-2p}-\frac{2 -p}{4}\geq \frac{1}{2}$$
	as $p\in(1,2]$, and $\bm{w}^0=\bm{w}^1$.
	
	\noindent	Combing the bounds \eqref{th3-b1} and \eqref{th3-b2}, we have
	\begin{align*}
		\frac{1}{T}\sum_{t=1}^T\EE\|\nabla f(\bm{w}^{t})\|&\leq \frac{L\sigma^{\frac{3}{8p}}(f(\bm{w}^{1})-\min f)+\frac{H}{L^2}\sigma^{\frac{4p-3}{4p}}+4\Big(\frac{B}{L}+1\Big)^{\frac{2 -p}{2}}(\sigma^{\frac{1}{5}}+\sigma^{\frac{1}{2}})+\sigma^{\frac{3}{8p}}}{T^{\frac{2p-2}{4p-1}}} \\
		&\quad+\frac{\|\bm{\epsilon}^{0}\|\sigma^{\frac{1}{2}}}{T^{\frac{2p-1}{4p-1}}}+\frac{L(f(\bm{w}^{0})-\min f)+1}{ \sqrt{T}} +\frac{4\Big(\frac{B}{L}+1\Big)^{\frac{2 -p}{2}}}{\sqrt{T}}+\frac{2\|\bm{\epsilon}^{0}\|}{T}.
	\end{align*}
	
	\section{Proofs of Technical Lemmas}
	\subsection{Proof of Lemma \ref{lemma-c}}
	From the scheme of the algorithms, we have
	\begin{align*}
		\|\bm{w}^{t+1}-\bm{w}^0\|=\Big\|\bm{w}^{t}-\gamma\frac{\bm{m}^t}{\|\bm{m}^t\|}-\bm{w}^0\Big\|\leq \gamma +\|\bm{w}^{t}-\bm{w}^{0}\|.
	\end{align*}
	For any $t\in \mathbb{Z}^+$, as $1 \leq t\leq T$, it holds
	\begin{align}\label{xian}
		\|\bm{w}^{t}-\bm{w}^{0}\|\leq \gamma T.
	\end{align}
	The Lipschitz continuity of the gradient indicates
	\begin{align*}
		\|\nabla f(\bm{w}^{t})\|&\leq \|\nabla f(\bm{w}^{0})\|+\|\nabla f(\bm{w}^{t})-\nabla f(\bm{w}^{0})\|\\
		&\leq \|\nabla f(\bm{w}^{0})\|+L\|\bm{w}^{t}-\bm{w}^{0}\|\leq \|\nabla f(\bm{w}^{0})\|+L\gamma T
	\end{align*}
	as $1\leq t\leq T$. If $h\geq 2(\|\nabla f(\bm{w}^{0})\|+L\gamma T)$, we then have $\|\nabla f(\bm{w}^{t})\|\leq \frac{h}{2}$. According to [Lemma 5, \citep{liu2023breaking}], we further get the following result
	\begin{align*}
		\begin{aligned}
			\EE\|\Ch_{h}(\bm{g}^t)-\EE\Ch_{h}(\bm{g}^t)\|^2&\leq \EE\|\Ch_{h}(\bm{g}^t)\|^2\leq 10\sigma^{p}h^{2-p}, \\
			\|\EE\Ch_{h}(\bm{g}^t)-\nabla f(\bm{w}^t)\|^2&\leq 4\sigma^{2p}h^{-2(p-1)}.
		\end{aligned}
	\end{align*}
	\subsection{Proof of Lemma \ref{lemma0}}
	Noticing that property \eqref{xian} still holds,
	the individual Lipschitz continuity of the gradient then yields
	\begin{align*}
		\|\nabla f(\bm{w}^{t};\xi^t)\|&\leq \|\nabla f(\bm{w}^{0};\xi^t)\|+\|\nabla f(\bm{w}^{t};\xi^t)-\nabla f(\bm{w}^{0};\xi^t)\|\\
		&\leq \|\nabla f(\bm{w}^{0};\xi^t)\|+L\|\bm{w}^{t}-\bm{w}^{0}\|\leq B+L\gamma T.
	\end{align*}
	Noticing $\|\nabla f(\bm{w}^{t})\|=\|\EE_{\xi^t\sim\mathcal{D}}\nabla f(\bm{w}^{t};\xi^t)\|\leq \EE_{\xi^t\sim\mathcal{D}}\|\nabla f(\bm{w}^{t};\xi^t)\|$, we then derive
	$$\|\nabla f(\bm{w}^{t})\|\leq B+L\gamma T$$
	as $1\leq t\leq T$.
	Direct computations give us
	\begin{align*}
		\|\nabla f(\bm{w}^t;\xi^t)-\nabla f(\bm{w}^t)\|^2&=\|\nabla f(\bm{w}^t;\xi^t)-\nabla f(\bm{w}^t)\|^{2-p}\cdot\|\nabla f(\bm{w}^t;\xi^t)-\nabla f(\bm{w}^t)\|^p\\
		&\leq (\|\nabla f(\bm{w}^t;\xi^t)\|+\|\nabla f(\bm{w}^t)\|)^{2-p}\cdot\|\nabla f(\bm{w}^t;\xi^t)-\nabla f(\bm{w}^t)\|^p\\
		&\leq (2(B+L\gamma T))^{2-p}\cdot\|\nabla f(\bm{w}^t;\xi^t)-\nabla f(\bm{w}^t)\|^p\\
		&\leq 4(B+L\gamma T)^{2-p}\|\nabla f(\bm{w}^t;\xi^t)-\nabla f(\bm{w}^t)\|^p,
	\end{align*}
	where $0\leq 2-p\leq 2$. Taking expectations of both sides, we can get
	\begin{align*}
		\EE_{\xi^t\sim\mathcal{D}}\|\nabla f(\bm{w}^t;\xi^t)-\nabla f(\bm{w}^t)\|^2&\leq 4(B+L\gamma T)^{2-p}\EE_{\xi^t\sim\mathcal{D}}\|\nabla f(\bm{w}^t;\xi^t)-\nabla f(\bm{w}^t)\|^p\\
		&\leq 4(B+L\gamma T)^{2-p}\sigma^p.
	\end{align*}
	\subsection{Proof of Lemma \ref{lemma1}}
	The proof of this lemma can be found in \citep{cutkosky2020momentum,zhao2021convergence} and is presented here for completeness.
	The gradient Lipschitz property can derive the following result
	\begin{align*}
		f(\bm{w}^{\ddag})-f(\bm{w}^{\dag})&\leq \langle\nabla f(\bm{w}^{\dag}),\bm{w}^{\ddag}-\bm{w}^{\dag}\rangle+\frac{L}{2}\|\bm{w}^{\ddag}-\bm{w}^{\dag}\|^2\\
		&\leq-\gamma \langle\nabla f(\bm{w}^{\dag}),\frac{\bm{m}}{\|\bm{m}\|}\rangle+\frac{L\gamma^2}{2}\\
		&\overset{a)}{=}-\gamma \|\bm{m}\|+\gamma \langle\bm{m}-\nabla f(\bm{w}^{\dag}),\frac{\bm{m}}{\|\bm{m}\|}\rangle+\frac{L\gamma^2}{2}\\
		&\overset{b)}{\leq}-\gamma \|\nabla f(\bm{w}^{\dag})\|+2\gamma\|\bm{m}-\nabla f(\bm{w}^{\dag})\|+\frac{L\gamma^2}{2},
	\end{align*}
	where $a)$ comes from the following fact
	\begin{align*}
		-\langle\nabla f(\bm{w}^{\dag}),\frac{\bm{m}}{\|\bm{m}\|}\rangle&=-\langle\bm{m},\frac{\bm{m}}{\|\bm{m}\|}\rangle-\langle\nabla f(\bm{w}^{\dag})-\bm{m},\frac{\bm{m}}{\|\bm{m}\|}\rangle\\
		&=-\|\bm{m}\|+\langle\bm{m}-\nabla f(\bm{w}^{\dag}),\frac{\bm{m}}{\|\bm{m}\|}\rangle\\
		&\leq -\|\bm{m}\|+\|\bm{m}-\nabla f(\bm{w}^{\dag})\|,
	\end{align*}
	and $b)$ is because
	$$- \|\bm{m}\|\leq -\|\nabla f(\bm{w}^{\dag})\|+\|\nabla f(\bm{w}^{\dag})-\bm{m}\|.$$

	\vskip 0.2in
	\bibliography{ref}
\end{document}